\documentclass{article}

\usepackage{times}
\usepackage{graphicx} 
\usepackage{subfigure} 
\usepackage{pifont}

\usepackage{natbib}

\usepackage{algorithm}
\usepackage{algorithmic}
\usepackage{multirow}
\usepackage{cancel}
\usepackage{empheq}

\usepackage{bm}
\usepackage{amssymb,amsmath,amsfonts,graphicx,amsthm}
\newtheorem{theorem}{Theorem}
\newtheorem{assumption}{Assumption}

\usepackage[accepted]{icml2016}

\icmltitlerunning{Deep Robust Kalman Filter}

\begin{document} 

\twocolumn[
\icmltitle{Deep Robust Kalman Filter}

\icmlauthor{Shirli Di-Castro Shashua}{shirlidi@tx.technion.ac.il}
\icmladdress{Technion - Israel Institute of Technology, Haifa, Israel}
\icmlauthor{Shie Mannor}{shie@ee.technion.ac.il}
\icmladdress{Technion - Israel Institute of Technology, Haifa, Israel}

\icmlkeywords{Robust MDP, Kalman filter, EKF}

\vskip 0.3in
]

\begin{abstract} 
	A Robust Markov Decision Process (RMDP) is a sequential decision making model that accounts for uncertainty in the parameters of dynamic systems. This uncertainty introduces difficulties in learning an optimal policy, especially for environments with large state spaces. We propose two algorithms, RTD-DQN and Deep-RoK, for solving large-scale RMDPs using  nonlinear approximation schemes such as deep neural networks. The RTD-DQN algorithm incorporates the robust Bellman temporal difference error into a robust loss function, yielding robust policies for the agent. The Deep-RoK algorithm is a robust Bayesian method, based on the Extended Kalman Filter (EKF), that accounts for both the uncertainty in the weights of the approximated value function and the uncertainty in the transition probabilities, improving the robustness of the agent. We provide theoretical results for our approach and test the proposed algorithms on a continuous state domain. 
\end{abstract} 

\section{Introduction }
\label{introduction}
Sequential decision making in stochastic environments are often modeled as Markov Decision Processes (MDPs) in order to optimize a policy that achieves maximal expected accumulated reward \cite{puterman2014markov, bertsekas1996neuro}. Given the MDP model parameters, namely the transition probabilities and the reward function, the aim of an agent is to find the optimal policy. In many cases the true MDP model is unknown in advance and its parameters are  estimated from data. The deviation of the estimated model from the true model may cause a degradation in the performance of the learned policy \cite{mannor2007bias}. This undesired behavior leads to the need for model-robust methods that consider a set of possible MDP models and find the optimal policy over this set.

\begin{figure}[t]
	\centering{
		\includegraphics[width=1.0\linewidth,height=0.5\textheight,keepaspectratio]{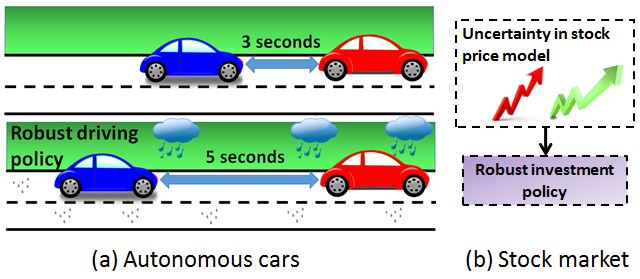}}
	\caption{Robustness in different domains.}
	\label{fig:robust_applications}
\end{figure}

Robustness is effective when environmental safety issues arise \cite{garcia2015comprehensive, lipton2016combating}. For example, in autonomous cars  (Figure \ref{fig:robust_applications}(a)) it is important to account for environmental uncertainties such as weather or road conditions. A robust driving policy should account for all of these uncertainties and must subsequently adjust the agent's driving behavior accordingly. Additional motivation for model-robust methods is when an agent seeks to optimize a {\it coherent risk measure} \cite{artzner1999coherent} or to follow a {\it risk-sensitive} policy \cite{petrik2011robust, chow2015risk}. For example, investing in stock markets (Figure  \ref{fig:robust_applications}(b)) requires defining a model for the dynamics of the stock prices. This model is based on historical data that may be noisy or insufficient which induces uncertainty in the model. A robust investing policy would account for the uncertain model to avoid dangerous decisions that can cause loss of large amount of money. Control tasks can benefit from robustness as well when state transitions depend on several parameters \cite{rajeswaran2016epopt}. A robust agent would consider different possible values for these parameters to ensure satisfactory performance in the real world. 

The robust-MDP (RMDP) framework \cite{nilim2005robust,iyengar2005robust} was developed  to account for uncertainties in the MDP model parameters when looking for an optimal policy. In this framework, each policy is associated with a {\it known} uncertainty set of transition probabilities. The optimal policy is the one that maximizes the {\it worst case} value function over the associated uncertainty set. When the state space is large, solving robust optimization problems can be a difficult task \cite{le2007robust}.

We are looking for a method for solving large-scale RMDPs, with {\it on-line nonlinear} value function approximation. Existing methods for solving RMDPs \cite{iyengar2005robust,nilim2005robust,tamar2014scaling,rajeswaran2016epopt} have several limitations such as linear approximation, off-line estimation and restrictive assumptions on the transition probabilities. We review these methods in Section \ref{Section:Related Work} and compare them to the  robust Bayesian approach we propose in this paper. 

We distinguish between two types of uncertainty: the {\it MDP model parameters uncertainty} refers to all the possible transition probability distributions of stepping from state $s$ to state $s'$ when taking an action $a$. We denote this set by $\mathcal{P}_{(\cdot|s,a)}$, and when it is clear from the context, we omit the subscript and use $\mathcal{P}$. The other type of uncertainty origins from the Bayesian assumption over the {\it weights} of the approximated value function. We denote the {\it weight uncertainty set} by $\Theta$. 

Inspired by the success of Deep Q-Network (DQN) agents \cite{mnih2013playing} in estimating large-scale nonlinear value functions, we propose the {\bf R}obust {\bf T}emporal {\bf D}ifference {\bf DQN} ({\bf RTD-DQN}) algorithm which replaces the nominal Bellman Temporal Difference (TD) error involved in the optimized objective function with the robust Bellman TD error. We show that this algorithm captures the MDP model uncertainty and improves the robustness of the agent.

The Kalman filter \cite{kalman1960new} and its variant for nonlinear approximations, the Extended Kalman filter (EKF) \cite{anderson1979optimal, gelb1974applied}, are used for on-line tracking and for estimating states in dynamic environments through indirect observations. These methods have been successfully applied to numerous control dynamic systems such as navigation and tracking targets. The Kalman filter can be also used for weights estimation of approximation functions, where the weights constitute the states of dynamic systems. We suggest to extend the RTD-DQN algorithm by using a Bayesian approach such as EKF to account for the uncertainty in the weights of the value function approximation, in addition to the uncertainty of the transition probabilities. We present this approach in the {\bf Deep} {\bf Ro}bust {\bf K}alman filter ({\bf Deep-RoK}) algorithm. 

\begin{figure}[t]
	\centering{
		\includegraphics[width=0.8\linewidth,height=0.7\textheight,keepaspectratio]{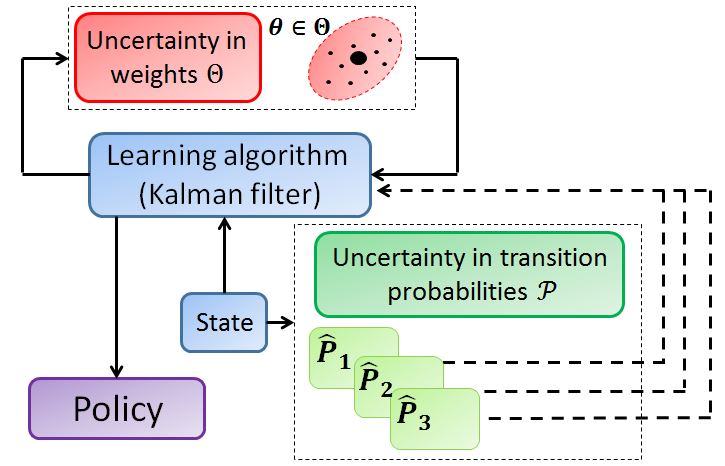}} 
	\caption{The robust Bayesian approach in Deep-RoK.}
	\label{fig:robust_approach}
\end{figure}

This approach may be surprising. How are the weights of the value function  related to the MDP model parameters? To answer this question we refer to the work of \citet{mannor2007bias}. When estimating the MDP model parameters, the potential for error in the estimates, i.e., the uncertainty in the MDP model parameters, introduces variance in the estimates of the value function, governed by the value function weights. In Figure \ref{fig:robust_approach} we illustrate our robust Bayesian approach. The EKF serves as a Bayesian learning algorithm that receives the new information from the transition probabilities uncertainty set $\mathcal{P}$ and propagates it into the weights uncertainty set $\Theta$. This approach provides a robust and efficient estimation as we demonstrate in the experiments.  

Our contributions are: (1) A Bayesian approach for on-line and nonlinear approximation of the value function in RMDPs. We connect the robust Bellman TD error to the EKF updates to achieve robust policies in RMDPs;  
(2) We propose two algorithms, RTD-DQN and Deep-RoK, to solve large scale RMDPs; (3) We provide theoretical guarantees for our proposed methods; (4) We demonstrate the performance of our two algorithms on a continuous state domain.

\section{Related Work}
\label{Section:Related Work}

\begin{table*}[t]
	\small
	\caption{Comparison of different approaches to Deep-RoK and RTD-DQN}
	\label{Table: ComparisonRMDPs}
	\begin{center}
		\begin{tabular}{ | c | c | c | c | c | c | c | c | }
			\hline
			& State  & \multirow{3}{*}{On-line}   &  Nonlinear    & Uncertainty in  & Uncertainty &   Kalman & RL\\ 
			& space & &  approximation & MDP model  & in weights & Filter  & method\\ 
			& Scalability & & & (robust) & (Bayesian) & & \\ \hline \hline
			\multirow{2}{*}{Deep-RoK (This paper)}  & \multirow{2}{*}{\checkmark} & \multirow{2}{*}{\checkmark} & \multirow{2}{*}{\checkmark} & \multirow{2}{*}{\checkmark} & \multirow{2}{*}{\checkmark} & \checkmark  & EKF for\\ 
			&   &  &  &  &  & (EKF) & Deep Q-learning\\ \hline
			RTD-DQN (This paper) & \checkmark  & \checkmark &  \checkmark & \checkmark & $\times$ & $\times$ & Deep Q-learning \\ \hline
			\cite{iyengar2005robust} & $\times$ & $\times$ & $\times$ &  \checkmark & $\times$ & $\times$ & DP \\ \hline
			\cite{nilim2005robust} & $\times$ & $\times$ & $\times$ &  \checkmark & $\times$ & $\times$  & DP \\ \hline
			\cite{tamar2014scaling} & \checkmark & $\times$ & $\times$ &  \checkmark & $\times$ & $\times$ & ADP \\ \hline
			\cite{rajeswaran2016epopt} & \checkmark  & $\times$ & \checkmark &  \checkmark & $\times$ & $\times$  & Policy gradient\\ \hline
			\cite{blundell2015weight} & \checkmark  & \checkmark &   \checkmark & $\times$ &  \checkmark & $\times$ & $\times$ \\ \hline
			\cite{li2016learning} & \checkmark  & \checkmark &  \checkmark & $\times$ &  \checkmark & $\times$ & $\times$ \\ \hline
			\multirow{2}{*}{ \cite{geist2010kalman}}  & \multirow{2}{*}{\checkmark}  & \multirow{2}{*}{$\times$} &  \multirow{2}{*}{\checkmark} & \multirow{2}{*}{$\times$} &  \multirow{2}{*}{\checkmark} & \checkmark  & UKF for\\
			 &  &  &  &  &   & (UKF) & Q-learning\\ \hline
			\cite{singhal1988training} & \checkmark  & $\times$ &  \checkmark & $\times$ &  \checkmark & \checkmark (EKF)& $\times$ \\ \hline
			 \hline
		\end{tabular}
	\end{center}
\end{table*}

Our paper is related to several areas of research, namely RMDPs, Deep Q-learning networks, EKF and Bayesian approach for weight uncertainty in Neural Networks (NNs). Our work is the first to solve RMDPs while combining scalability to large state spaces, on-line estimation, nonlinear Q-function approximations, robustness to uncertainty in the transition probabilities and a Bayesian approach (EKF) to account for the uncertainty in the approximation weights. In Table \ref{Table: ComparisonRMDPs} we compare between different approaches and our proposed algorithms,  Deep-RoK and RTD-DQN.  

\citet{tamar2014scaling} used approximate dynamic programming (ADP) method with linear value function approximation. Their convergence analysis is based on a restrictive assumption over the transitions of the exploration policy and the (uncertain) transitions of the policy under evaluation. Our work does not rely on such assumptions which facilitates the convergence analysis of our proposed algorithms. 

The use of Kalman filters to solve reinforcement learning (RL) problems was proposed by \citet{geist2010kalman}. Their formulation, called Kalman Temporal Difference (KTD), serves as the base for our formulation for the algorithms we propose. We introduce here several differences between their work and ours: (1) We re-formulate the observation function such that the observation of the agent at time $t$ is the target label, meaning the sum of the immediate reward and the discounted next state optimal Q-function. With this formulation, the observation function is simply the Q-function of the current state and action; (2) They used the nominal Bellman TD error, while we are using the robust version of it; (3) We use the Extended Kalman filter as opposed to their use of the Unscented Kalman filter (UKF) to approximate nonlinear functions \cite{julier1997new, wan2000unscented}. In our formulation, the observation function is differential, allowing us to use first order Taylor expansion linearization as used in the EKF and in gradient descent optimization methods. The UKF has shown superior performance in some applications \cite{st2004comparison, van2004sigma}, however, its computational cost is much greater than the computational cost of the EKF, due to its requirement of sampling the weights in each training step for number of times equals to the double of the weights dimension. Moreover, it requires to evaluate the observation function at these samples at every training step. Unfortunately, this is not tractable in deep NNs where the weights might be high-dimensional. 

\section{Background}
In this section we describe formulations from different fields, towards combining them into our formulation for solving large scale RMDPs in Section \ref{section:Solving Large-Scale RMDPs}.

\subsection{Robust Markov Decision Processes (RMDPs)}
\label{Robust Markov Decision Processes}
An RMDP \citep{iyengar2005robust, nilim2005robust} is a tuple $\{\mathcal{S}, \mathcal{A}, \mathcal{P}_{(\cdot | s,a)}, R, \gamma\}$ where $\mathcal{S}$ is a finite set of states, $\mathcal{A}$ is a finite set of actions, $R: \mathcal{S} \times \mathcal{A} \rightarrow \mathbb{R}$ is a deterministic and bounded reward function, $\gamma$ is a discount factor and  $\mathcal{P}_{(\cdot | s,a)}$ is a probability measure over $(s,a)$ which denotes the uncertainty set of the transition probabilities for each state $s$ and action $a$. At each discrete time step $t$ the system stochastically steps from state $s_t \in \mathcal{S}$ to state $s_{t+1} \in \mathcal{S}$ by taking an action $a_t \in \mathcal{A}$.  Each transition $(s_t, a_t, s_{t+1})$ is associated with an immediate reward $r_t(s_t, a_t)$. The agent chooses the actions according to a policy $\pi: \mathcal{S} \rightarrow \mathcal{A}$ that maps each state to a probability distribution over the actions set. The transitions in the system are according to the probability distribution $P(s_{t+1}|s_t, a_t)$ which is assumed to lie in a {\em known} uncertainty set $P \in \mathcal{P}_{(\cdot|s_t, a_t)}$. 
 
The Q-function of state-action pair $(s,a)$ under policy $\pi$ and state transition model $P$ represents the expected sum of discounted returns when starting from $(s,a)$ and executing policy $\pi$: $Q^{\pi, P}(s, a) = \mathbb{E}^{\pi, P}\big[ \sum_{t=0}^{\infty} \gamma^t r_t(s_t, a_t)  | s_0 = s, a_0 = a \big]$, where $\mathbb{E}^{\pi, P}$ denotes the expectation w.r.t. the state-action distribution induced by the transitions $P$ and the policy $\pi$. In RMDPs, we are interested in finding the policy that maximizes the {\em worst case} Q-function: $Q^{\pi}(s, a) = \inf_{P \in \mathcal{P}} Q^{\pi, P}(s, a)$.
The optimal robust Q-function is then the unique solution of the robust Bellman recursion:
\begin{align}
\label{eq:robustQfunction}
Q^*(s, a) & = \sup_\pi \{\inf_{P \in \mathcal{P}} Q^{\pi, P}(s, a) \}\\
\nonumber & = r(s, a) +  \gamma \inf_{P \in \mathcal{P}} \mathbb{E}^P [ \max_{a'} Q^* (s', a')|s, a ],
\end{align} 
where $s'$ is the successive state when taking an action $a$ in state $s$. \citet{iyengar2005robust} showed that the agent can be restricted to stationary deterministic Markov policies without affecting the achievable robust reward. In this paper we focus on the $\epsilon$-greedy exploration strategy, where the agent takes a random action with probability $\epsilon$, and follows the optimal policy with probability $1-\epsilon$. 

The solution of the minimization problem in Equation  (\ref{eq:robustQfunction}) may be computationally demanding. Fortunately, there are some families of sets $\mathcal{P}$ for which the solution is tractable. Popular uncertainty sets, presented by \citet{iyengar2005robust} and \citet{nilim2005robust} are constructed from approximations of confidence regions associated with probability density estimation. This choice seems natural when the uncertainty is due to statistical errors when estimating the states transition probabilities using historical data. 

\subsection{Deep Q-learning} 
Q-learning \cite{watkins1992q} is a TD method that aims at directly finding the optimal policy by updating the Q-function with the optimal greedy policy $a^* = \arg\max_{a'} Q^*(s, a')$ \cite{sutton1988learning}. Therefore, learning the optimal policy can be reduced to learning the optimal Q-function. In many RL problems the state space is large, thus the Q-function is typically approximated by parametric models $Q(s, a; \boldsymbol{\theta})$ where $\boldsymbol{\theta}$ denotes the weights of the approximation function. 

In Deep Q-learning \cite{mnih2013playing}, the agent improves the Q-function (and, in turn, the greedy policy) by minimizing at each time step $t$ the squared {\it nominal Bellman TD error} (nBTDe)  $\delta^{\text{nominal}}_{t, \boldsymbol{\theta}'}$:
\begin{equation}
\label{DQN objective}
L^{\text{nominal}}_t(\boldsymbol{\theta}_t) = \frac{1}{2} \mathbb{E}_{o_t \sim p(\cdot)}[\big( \delta^{\text{nominal}}_{t, \boldsymbol{\theta}'} (\boldsymbol{\theta}_t, o_t) \big)^2],
\end{equation}
where
\begin{equation}
\label{eq: Bellman TD}
\delta^{\text{nominal}}_{t, \boldsymbol{\theta}'} (\boldsymbol{\theta}_t, o_t) \triangleq y^{\text{nominal}}_{t, \boldsymbol{\theta}'} (o_t) - Q(s_t, a_t; \boldsymbol{\theta}_t).
\end{equation}
Here, $y^{\text{nominal}}_{t, \boldsymbol{\theta}'}$ is the {\it nominal target label} and is defined as:
\begin{equation}
\label{target label DQN}
y^{\text{nominal}}_{t, \boldsymbol{\theta}'} (o_t) \triangleq  r_t + \gamma \max_{a'} Q(s_{t+1}, a'; \boldsymbol{\theta}').
\end{equation} 
The weight $\boldsymbol{\theta}'$ is a fixed set of weights, normally called the {\it  target network}. It is composed of a more stable periodic copy of the trained weights. We denote by $p(\cdot)$ the joint distribution of experiences under the current policy. The observation in each time step $o_t = (s_t, a_t, r_t, s_{t+1})$ is typically stored in an experience replay $\mathcal{O} = o_1, \ldots, o_N$. 

Traditionally, the Q-function is trained by stochastic gradient descent, estimating the loss on each experience as it is encountered, yielding the update:
\begin{align}
\label{eq:GradientDescent}
\nonumber \boldsymbol{\theta}_{t+1}  \leftarrow  \boldsymbol{\theta}_t + \alpha \mathbb{E}_{o_t \sim p(\cdot)} \big[ & \big(y^{\text{nominal}}_{t, \boldsymbol{\theta}'} (o_t) - Q(s_t, a_t; \boldsymbol{\theta}_t)\big)\\ 
& \nabla_{\boldsymbol{\theta}_t} Q(s_t, a_t; \boldsymbol{\theta}_t) \big], 
\end{align}
where $\alpha$ is the learning rate. 

\subsection{Deep Q-learning: A Bayesian Perspective}
The weights $\boldsymbol{\theta}$ can be learned by maximum likelihood estimation (MLE) using stochastic gradient descent methods: $\boldsymbol{\theta}^{\text{MLE}} = \arg\max_{\boldsymbol{\theta}} \log p(\mathcal{O}|\boldsymbol{\theta})$.
A Bayesian approach uses Bayes rule and suggests adding  prior knowledge over the weights $p(\boldsymbol{\theta})$ to calculate the maximum  a-posteriori (MAP) estimator:
\begin{align}
\label{MAP}
\nonumber \boldsymbol{\theta}^{\text{MAP}} & = \arg\max_{\boldsymbol{\theta}} \log p(\boldsymbol{\theta}|\mathcal{O})\\
& = \arg\max_{\boldsymbol{\theta}} \log p(\mathcal{O}|\boldsymbol{\theta}) + \log p(\boldsymbol{\theta}).
\end{align}
Placing a prior introduces regularization to the network. We will address the advantages of this regularization in Section \ref{section:Solving Large-Scale RMDPs}.

\subsection{Extended Kalman Filter (EKF)}
\label{Section:EKF}
In this section we briefly outline the Extended Kalman filter \cite{anderson1979optimal, gelb1974applied}. The EKF is a standard technique for estimating the state of a nonlinear dynamic system or for learning the weights of a nonlinear approximation function. 
In this paper we will focus on its latter role, meaning estimating $\boldsymbol{\theta}$. The EKF considers the following model:
\begin{equation}
\label{eq:Extended-Kalman}
\begin{cases}
\boldsymbol{\theta}_t = \boldsymbol{\theta}_{t-1} + {\bf v}_t\\
o_t = h(\boldsymbol{\theta}_t) + n_t
\end{cases},
\end{equation}
where $\boldsymbol{\theta}_t$ are the weights evaluated at time $t$, $o_t$ is the observation at time $t$ and $h(\boldsymbol{\theta}_t)$ is a nonlinear observation function. ${\bf v}_t$ is the evolution noise, $n_t$ is the observation noise, both modeled as additive and white noises with covariance ${\bf P}_{{\bf v}_t}$ and variance $P_{n_t}$ respectively. As seen in the model presented in Equation (\ref{eq:Extended-Kalman}) the EKF treats the weights $\boldsymbol{\theta}_t$ as random variables, similarly to the Bayesian approach. According to this perspective, the weights belong to an uncertainty set $\Theta$ governed by the mean and covariance of the weights distribution. 

The estimation at time t, denoted as $\boldsymbol{\hat{\theta}}_{t|\cdot}$ is the conditional expectation of the weights with respect to the observed data. The EKF formulation distinguishes between estimates that are based on observations up to time $t$, $\boldsymbol{\hat{\theta}}_{t|t}  \triangleq \mathbb{E} [ \boldsymbol{\theta}_t | o_{1:t}]$, and observations up to time $t-1$, $\boldsymbol{\hat{\theta}}_{t|t-1}  \triangleq \mathbb{E} [ \boldsymbol{\theta}_t | o_{1:t-1}]  = \boldsymbol{\hat{\theta}}_{t-1|t-1}$. The {\it weights errors} are defined by: $\boldsymbol{\tilde{\theta}}_{t|t}  \triangleq \boldsymbol{\theta}_t - \boldsymbol{\hat{\theta}}_{t|t}$ and $\boldsymbol{\tilde{\theta}}_{t|t-1} \triangleq \boldsymbol{\theta}_t - \boldsymbol{\hat{\theta}}_{t|t-1}$.
 
The conditional {\it error covariances} are given by: 
\begin{align*}
{\bf P}_{t|t}  \triangleq \mathbb{E} \big[  \boldsymbol{\tilde{\theta}}_{t|t}  \boldsymbol{\tilde{\theta}}_{t|t}^\top | o_{1:t}\big], \ 
 {\bf P}_{t|t-1} & \triangleq \mathbb{E} \big[  \boldsymbol{\tilde{\theta}}_{t|t-1}  \boldsymbol{\tilde{\theta}}_{t|t-1}^\top | o_{1:t-1}\big] \\
 & = {\bf P}_{t-1|t-1} + {\bf P}_{{\bf v}_t}.
\end{align*}
EKF considers several statistics of interest at each time step:
{\it The prediction of the observation function}, {\it the observation innovation}, {\it the covariance between the weights error and the innovation}, {\it the covariance of the innovation}  and the {\it Kalman gain} are defined respectively in Equations (\ref{eq:The prediction of the observation}) - (\ref{eq:Kal_gain}):
\begin{align}
\label{eq:The prediction of the observation} & \hat{o}_{t|t-1} \triangleq  \mathbb{E}[h(\boldsymbol{\theta}_t)|o_{1:t-1}],\\
\label{eq:The observation innovation} & \tilde{o}_{t|t-1} \triangleq h(\boldsymbol{\theta}_t) - \hat{o}_{t|t-1}, \\
\label{eq:The covariance between the weights error and the innovation} & {\bf P}_{\boldsymbol{\tilde{\theta}}_t,\tilde{o}_t}  \triangleq \mathbb{E}[ \boldsymbol{\tilde{\theta}}_{t|t-1} \tilde{o}_{t|t-1} |o_{1:t-1}],\\	
\label{eq:The covariance of the innovation} & P_{\tilde{o}_t}  \triangleq \mathbb{E}[( \tilde{o}_{t|t-1})^2|o_{1:t-1}] + P_{n_t},\\
\label{eq:Kal_gain} & {\bf K}_t \triangleq {\bf P}_{\boldsymbol{\tilde{\theta}}_t,\tilde{o}_t} P_{\tilde{o}_t}^{-1}.
\end{align}
The above statistics serve for the update of the weights and the error covariance:
\[\begin{cases}
\boldsymbol{\hat{\theta}}_{t|t} = \boldsymbol{\hat{\theta}}_{t|t-1} +  \mathbb{E}_{o_t \sim p(\cdot)} \big[ {\bf K}_t \big( o_t - h(\boldsymbol{\hat{\theta}}_{t|t-1}) \big) \big],\\
{\bf P}_{t|t} = {\bf P}_{t|t-1} - {\bf K}_t P_{\tilde{o}_t} {\bf K}_t^\top.
\end{cases}\]

\section{Solving Large-Scale RMDPs}
\label{section:Solving Large-Scale RMDPs}
In this section we explain how to combine EKF as a Bayesian method with Deep Q-learning and the robust formulation to solve large scale RMDPs with uncertainty in the transition probabilities. 

\subsection{Robust Temporal Difference Deep Q-Network (RTD-DQN)}
Our first step in solving large scale RMDPs with on-line nonlinear approximations is to change the nBTDe presented in Equation (\ref{target label DQN}) with the {\it robust Bellman TD error} (rBTDe) $\delta^{\text{robust}}_{t, \boldsymbol{\theta}'}$ and minimize the following objective function at each time step:
$$ L_t^{\text{robust}} (\boldsymbol{\theta}_t)  = \frac{1}{2} \mathbb{E}_{o_t \sim p(\cdot)} [\big( \delta^{\text{robust}}_{t, \boldsymbol{\theta}'}(\boldsymbol{\theta}_t, o_t) \big)^2], $$
where 
\begin{equation}
\label{eq:robust Bellman TD error}
\delta_{t, \boldsymbol{\theta}'}^{\text{robust}}( \boldsymbol{\theta}_t, o_t) = y_{t, \boldsymbol{\theta}'}^{\text{robust}} (o_t) - Q (s_t, a_t; \boldsymbol{\theta}_t).
\end{equation}
Here, $y_{t, \boldsymbol{\theta}'}^{\text{robust}}$ is the {\it robust target label}:
\begin{equation}
\label{eq:target_robust_label}
y_{t, \boldsymbol{\theta}'}^{\text{robust}}(o_t) = r_t + \gamma \min_{p \in \mathcal{P}} \sum_{s'\in \tilde{\mathcal{S}}} p(s'|s_t, a_t) \max_{a'} Q(s', a'; \boldsymbol{\theta}').
\end{equation}
The set $\tilde{\mathcal{S}}$ is the set of all possible next states from state $s_t$ when taking action $a_t$, and it is drawn from the uncertainty set  $\mathcal{P}_{(\cdot| s_t, a_t)}$. Note that $y_{t, \boldsymbol{\theta}'}^{\text{robust}}$ is a variation of the robust Bellman function for the optimal Q-function presented in Equation (\ref{eq:robustQfunction}). It looks for worst case transitions that may reduce the value of the expected Q-function, and sets the robust target label value according to the minimal expectation. In return, the agent that receives these robust target labels, learns how to act optimally over these transitions.

This approach for solving RMDPs is presented in Algorithm \ref{alg:RTD-DQN}, RTD-DQN. It is based on the DQN algorithm \cite{mnih2013playing} but incorporates the robust target label instead of the nominal one. 
RTD-DQN initializes the weights $\boldsymbol{\hat{\theta}}_{0|0}$ with small random values and holds an experience reply $\mathcal{O}$ with finite capacity $N$. The environment is initialized at the beginning of each episode. We denote the estimation process with subscript $t$, while the observations at each episode are denoted with subscript $i$. During an episode the agent observes the uncertainty set $\mathcal{P}_{(\cdot| s_i, a_i)}$ for the set of next possible states $\tilde{\mathcal{S}}_{(\cdot| s_i, a_i)}$. For each mini-batch sample $o_j$, the agent computes  $y_{j, \boldsymbol{\theta}'}^{\text{robust}}(o_j)$  (\ref{eq:target_robust_label}) and updates the weights according to the gradient descent step  (\ref{eq:GradientDescent}) with $y_{j, \boldsymbol{\theta}'}^{\text{robust}}$ replacing $y_{j, \boldsymbol{\theta}'}^{\text{nominal}}$. Note that the environment transitions to state $s_{i+1}$ that is drawn from the true unknown MDP model parameter $P$. However, actions taken by the agent adhere the robust policy, based on the robust target labels. The output of the RTD-DQN algorithm is the MLE weights estimator $\boldsymbol{\hat{\theta}}^{MLE}_{t|t}$. 
The RTD-DQN algorithm incorporates the robust target label into the weights update, but it does not account for uncertainty in the weights. This uncertainty is important for robustness properties. In the next section we suggest to use EKF for this purpose.  
\begin{algorithm}[t]
	\caption{RTD-DQN}
	\label{alg:RTD-DQN}
	\begin{algorithmic}[1]
		\REQUIRE $\mathcal{P}$, $\gamma$, $\alpha$; {\bf Initialize}: $\mathcal{O}$, $\boldsymbol{\hat{\theta}}_{0|0}$, $t=0$.
		\FOR {episode $=1 \ldots, M$}
		\FOR {$i=1, \ldots$}
		\STATE With probability $\epsilon$ select a random action $a_i$, otherwise select $a_i = \arg\max_{a'} Q(s_i, a'; \boldsymbol{\hat{\theta}}_{t|t-1})$
		\STATE Observe $o_i = \{s_i, a_i, r_i, s_{i+1}, \tilde{S}_{(\cdot| s_i, a_i)}\}$  and store it in $\mathcal{O}$.
		\STATE Compute $y_{j, \boldsymbol{\theta}'}^{\text{robust}}(o_j)$ (\ref{eq:target_robust_label}) for random mini-batch  $\{o_j\}_{j=1}^k \in \mathcal{O}$.
		\STATE Compute $\nabla_{\boldsymbol{\theta}_t} Q(s_j, a_j; \boldsymbol{\hat{\theta}}_{t|t-1})$.
		\STATE Update the weights: \\
		$\begin{aligned}\boldsymbol{\hat{\theta}}_{t|t} = \ &  \boldsymbol{\hat{\theta}}_{t|t-1} +  \alpha \mathbb{E}_{o_j \sim \mathcal{O}} \big[ \big( y^{\text{robust}}_{t, \boldsymbol{\theta}'} (o_j)
		- \\
		& Q(s_j, a_j; \boldsymbol{\hat{\theta}}_{t|t-1}) \big) \nabla_{\boldsymbol{\theta}_t} Q(s_j, a_j; \boldsymbol{\hat{\theta}}_{t|t-1}) \big] \end{aligned}$
		\STATE $t = t+1$
		\ENDFOR
		\ENDFOR
		\ENSURE $\boldsymbol{\hat{\theta}_{t|t}}$
	\end{algorithmic}
\end{algorithm}

\subsection{EKF for Deep Q-learning}
\label{Section:EKF_Qlearning}
We propose to improve the performance of RTD-DQN by accounting for uncertainty in the weights in addition to the uncertainty in the transition probabilities. We suggest to use EKF for solving Deep Q-leaning networks. For this purpose, using the formulation for EKF presented in Section \ref{Section:EKF}, the observation at time $t$ is simply the nominal target label $y^{\text{nominal}}_{t, \boldsymbol{\theta}'} (o_t)$ (\ref{target label DQN}) and the observation function is the state-action Q-function, $h(\boldsymbol{\theta}_t) = Q(s_t, a_t; \boldsymbol{\theta}_t)$. With this formulation, the EKF model for DQN with Bayesian approach becomes:
\begin{equation}
\label{eq:Extended Kalman_DQN}
\begin{cases}
\boldsymbol{\theta}_t = \boldsymbol{\theta}_{t-1} + {\bf v}_t\\
r_t + \gamma \max_{a'} Q(s_{t+1}, a'; \boldsymbol{\theta}') = Q(s_t, a_t; \boldsymbol{\theta}_t) + n_t
\end{cases}.
\end{equation}
The EKF uses a first order Taylor series linearization for the observation function (Q-function):\\
\begin{equation*}
Q(s_t, a_t; \boldsymbol{\theta}_t) 
= Q(s_t, a_t; \boldsymbol{\hat{\theta}}) +  \big( \boldsymbol{\theta}_{t} - \boldsymbol{\hat{\theta}} \big)^\top \nabla_{\boldsymbol{\theta}_t} Q(s_t, a_t; \boldsymbol{\hat{\theta}}),
\end{equation*}
where $\boldsymbol{\hat{\theta}}$ is typically chosen to be the previous estimation of the weights at time $t-1$,  $\boldsymbol{\hat{\theta}}_{t|t-1}$. This linearization helps in computing the statistics of interest (see the supplementary material for more detailed derivations):
\begin{align}
\nonumber & \hat{y}_{t|t-1}(o_t)   = Q(s_t, a_t; \boldsymbol{\hat{\theta}}_{t|t-1}),\\
\nonumber & {\bf P}_{\boldsymbol{\tilde{\theta}}_t,\tilde{y}(o_t)}  = {\bf P}_{t|t-1} \nabla_{\boldsymbol{\theta}_t} Q(s_t, a_t; \boldsymbol{\hat{\theta}}_{t|t-1}), \\
\nonumber & P_{\tilde{y}(o_t)}   = \nabla_{\boldsymbol{\theta}_t} Q(s_t, a_t; \boldsymbol{\hat{\theta}}_{t|t-1})^\top {\bf P}_{t|t-1} \nabla_{\boldsymbol{\theta}_t} Q(s_t, a_t; \boldsymbol{\hat{\theta}}_{t|t-1}) \\
\label{eq:statistics2} & \quad \quad \quad \quad + P_{n_t}.
\end{align}
The Kalman gain then becomes:
\begin{align}
\label{eq:Kalman_gain}
{\bf K}_t & = {\bf P}_{t|t-1} \nabla_{\boldsymbol{\theta}_t} Q(s_t, a_t; \boldsymbol{\hat{\theta}}_{t|t-1}) \\
\nonumber & \Big( \nabla_{\boldsymbol{\theta}_t} Q(s_t, a_t; \boldsymbol{\hat{\theta}}_{t|t-1})^\top {\bf P}_{t|t-1} \nabla_{\boldsymbol{\theta}_t} Q(s_t, a_t; \boldsymbol{\hat{\theta}}_{t|t-1})  \\
\nonumber & + P_{n_t} \Big)^{-1}
\end{align}
and the update for the weights of the Q-function and the error covariance are:
\begin{align}
\label{eq:kalman update}
\begin{cases}
\boldsymbol{\hat{\theta}}^{\text{EKF}}_{t|t} = \boldsymbol{\hat{\theta}}_{t|t-1} +  \mathbb{E}_{o_t \sim p(\cdot)} \big[ {\bf K}_t \big( y^{\text{nominal}}_{t, \boldsymbol{\theta}'} (o_t) \\
\quad \quad \quad \quad \quad \quad \quad \quad \quad \quad  - Q_{}(s_t, a_t; \boldsymbol{\hat{\theta}}_{t|t-1}) \big) \big]\\
{\bf P}_{t|t} = {\bf P}_{t|t-1} - {\bf K}_t P_{\tilde{y}(o_t)} {\bf K}_t^\top.
\end{cases}
\end{align}
It is interesting to note that the Kalman gain ${\bf K}_t$ (\ref{eq:Kalman_gain}) can be interpreted as an adaptive learning rate  for each individual weight that implicitly incorporates the uncertainty of each weight. This approach resembles familiar stochastic gradient optimization methods such as Adagrad \cite{duchi2011adaptive}, AdaDelta \cite{zeiler2012adadelta}, RMSprop \cite{tieleman2012lecture}
and Adam \cite{kingma2014adam}, for different choices of ${\bf P}_{t|t-1}$ and $P_{n_t}$. We refer the reader to \citet{ruder2016overview}.

We now revisit the MAP estimator presented in Equation (\ref{MAP}). Given the observations gathered up to time $t$ (denoted as $o_{1:t}$) we can write:
\begin{align}
\label{eq:MAP_EKF}
\nonumber \boldsymbol{\theta}_t^{MAP} & = \arg\max_{\boldsymbol{\theta}_t} \log p(\boldsymbol{\theta}_t|o_{1:t})\\
& = \arg\max_{\boldsymbol{\theta}_t} \log p(o_{t}|\boldsymbol{\theta}_t) + \log  p(\boldsymbol{\theta}_t|o_{1:t-1}).
\end{align}
Here, instead of using the prior of the weights, we present an equivalent derivation for the posterior of the weights conditioned on $o_{1:t}$, based on the likelihood of a {\it single} observation $o_t$ and the posterior conditioned on $o_{1:t-1}$ \cite{van2004sigma}. When estimating using the EKF, it is common to make the following assumptions regarding the likelihood and the posterior:
\begin{assumption}
	\label{As:ConditionalIndependance}
	The likelihood $p(y_t(o_t)|\boldsymbol{\theta}_t)$ is assumed to be Gaussian: 
	 $y_t(o_t)|\boldsymbol{\theta}_t \sim \mathcal{N}(\mathbb{E}_{o_t \sim p(\cdot)} [Q(s_t, a_t; \boldsymbol{\theta}_t)], P_{n_t})$. 
\end{assumption}

\begin{assumption}
	\label{As:GaussianPosterior}
	The posterior distribution $p(\boldsymbol{\theta}_t|o_{1:t-1})$ is assumed to be Gaussian: $\boldsymbol{\theta}_t|o_{1:t-1} \sim \mathcal{N}(\boldsymbol{\hat{\theta}}_{t|t-1},{\bf P}_{t|t-1})$.
\end{assumption}
We use the notation $y_t(o_t)$ for a general target label (for example $y_{t, \boldsymbol{\theta}'}^{\text{nominal}}(o_t)$ or $y_{t, \boldsymbol{\theta}'}^{\text{robust}}(o_t)$) that serves as an observation in the EKF formulation. Based on the Gaussian assumptions, we can derive the following Theorem:
\begin{theorem}
	\label{theorem1}
	Under Assumptions \ref{As:ConditionalIndependance} and \ref{As:GaussianPosterior}, $\boldsymbol{\hat{\theta}}^{\text{EKF}}_{t|t}$  (\ref{eq:kalman update}) minimizes at each time step $t$ the following regularized objective function:
	\begin{align*}
	\nonumber L^{\text{EKF}}_t(\boldsymbol{\theta}_t) & =  \frac{1}{2P_{n_t}} \mathbb{E}_{o_t \sim p(\cdot)} [ \big(\delta^{\text{nominal}}_{t, \boldsymbol{\theta}'} (\boldsymbol{\theta}_t, o_t) \big)^2] \\
	& +  \frac{1}{2}(\boldsymbol{\theta}_t - \boldsymbol{\hat{\theta}}_{t|t-1})^\top {\bf P}_{t|t-1}^{-1} (\boldsymbol{\theta}_t - \boldsymbol{\hat{\theta}}_{t|t-1}),
	\end{align*}
	where $\boldsymbol{\hat{\theta}}^{\text{EKF}}_{t|t} \in \arg\min_{\boldsymbol{\theta}_t} L^{\text{EKF}}_t(\boldsymbol{\theta}_t)$.
\end{theorem}
The proof for Theorem \ref{theorem1} appears in the supplementary material. It is based on solving the maximization problem in Equation (\ref{eq:MAP_EKF}) using the EKF model (\ref{eq:Extended Kalman_DQN}) and the Gaussian assumptions in Assumptions \ref{As:ConditionalIndependance} and \ref{As:GaussianPosterior}.

Note that this objective function is a regularized version of the objective function in Equation (\ref{DQN objective}), where the weights are  weighted according to the error covariance matrix ${\bf P}_{t|t-1}$. The nBTDe $\delta^{\text{nominal}}_{t, \boldsymbol{\theta}'}$ is the same as in Equation (\ref{eq: Bellman TD}) and the nominal target label is the same as in Equation (\ref{target label DQN}). The observation noise variance $P_{n_t}$ can be interpreted as the regularization coefficient \cite{rivals1998recursive} and we can examine it from two points of view: (1) Looking at $P_{n_t}$ as the amount of confidence we have in the observations: if the observations are noisy, we should consider larger values for $P_{n_t}$. (2) Treating $P_{n_t}$ as a regularization coefficient: when observations are noisy we would like to put larger impact on the weights prior by increasing  $P_{n_t}$.

\subsection{Deep Robust Extended Kalman Filter (Deep-RoK)} 
\label{section: Deep-RoK}

\begin{algorithm}[t]
	\caption{Deep-RoK}
	\label{alg:Deep-RoK}
	\begin{algorithmic}[1]
		\REQUIRE ${\bf P}_{0|0}$, ${\bf P}_{{\bf v}_t}$,  $P_{n_t}$, $\gamma$, $\mathcal{P}$; {\bf Initialize}: $\mathcal{O}$, $\boldsymbol{\hat{\theta}}_{0|0}$, $t=0$
		\FOR {episode $=1 \ldots, M$}
		\FOR {$i=1, \ldots $}
		\STATE Set predictions: $\begin{cases}
		\boldsymbol{\hat{\theta}}_{t|t-1} = \boldsymbol{\hat{\theta}}_{t-1|t-1}\\
		{\bf P}_{t|t-1} = {\bf P}_{t-1|t-1} + {\bf P}_{{\bf v}_t}
		\end{cases}$.
		\STATE With probability $\epsilon$ select a random action $a_i$, otherwise select $a_i = \arg\max_{a'} Q(s_i, a'; \boldsymbol{\hat{\theta}}_{t|t-1})$
		\STATE Observe $o_i = \{s_i, a_i, r_i, s_{i+1}, \tilde{S}_{(\cdot | s_i, a_i)}\}$  and store it in $\mathcal{O}$.
		\STATE Compute $y_{j, \boldsymbol{\theta}'}^{\text{robust}}(o_j)$ (\ref{eq:target_robust_label}) for random mini-batch  $\{o_j\}_{j=1}^k \in \mathcal{O}$.
		\STATE Compute $\nabla_{\boldsymbol{\theta}_t} Q(s_j, a_j; \boldsymbol{\hat{\theta}}_{t|t-1})$.
		\STATE Compute $P_{\tilde{y}(o_j)}$  (\ref{eq:statistics2}) and  ${\bf K}_t$ (\ref{eq:Kalman_gain}).
		\STATE Update the weights and error covariance: $\begin{cases}
		\boldsymbol{\hat{\theta}}_{t|t} = \boldsymbol{\hat{\theta}}_{t|t-1} +  \mathbb{E}_{o_j \sim \mathcal{O}} \big[ {\bf K}_t \big( y^{\text{robust}}_{j, \boldsymbol{\theta}'} (o_j)\\
		\quad \quad \quad \quad \quad \quad \quad \quad  - Q(s_j, a_j; \boldsymbol{\hat{\theta}}_{t|t-1}) \big) \big]\\
		{\bf P}_{t|t} = {\bf P}_{t|t-1} - {\bf K}_t P_{\tilde{y}(o_j)} {\bf K}_t^\top
		\end{cases}$
		\STATE $t=t+1$
		\ENDFOR
		\ENDFOR
		\ENSURE $\boldsymbol{\hat{\theta}_{t|t}}$ and ${\bf P}_{t|t}$
	\end{algorithmic}
\end{algorithm}

We are now ready to combine the EKF as a Bayesian method with the rBTDe $\delta^{\text{robust}}_{t, \boldsymbol{\theta}'}$ (\ref{eq:robust Bellman TD error}). This approach incorporates uncertainty in the Q-function weights and allows propagation of the uncertainty in the transition probabilities, $\mathcal{P}$, into the uncertainty set of the weights, $\Theta$. This approach can be utilized to  solve large scale RMDPs with nonlinear approximation and in an on-line fashion. Practically we change the objective function presented in Equation (\ref{DQN objective}) by both adding regularization according to the EKF formulation and by replacing the nBTDe with the rBTDe $\delta^{\text{robust}}_{t, \boldsymbol{\theta}'}$. This results in the following Theorem:

\begin{theorem}
	\label{theorem2}
	Under Assumptions \ref{As:ConditionalIndependance} and \ref{As:GaussianPosterior}, $\boldsymbol{\hat{\theta}}^{\text{robust}}_{t|t}$  (\ref{eq:kalman update_robust}) minimizes at each time step $t$ the following regularized {\it robust objective function}:
	\begin{align*}
	\nonumber L_t^{\text{robust EKF}} (\boldsymbol{\theta}_t) & = \frac{1}{2P_{n_t}} \mathbb{E}_{o_t \sim p(\cdot)} [\big( \delta^{\text{robust}}_{t, \boldsymbol{\theta}'}(\boldsymbol{\theta}_t, o_t) \big)^2] \\
	\nonumber & + \frac{1}{2}(\boldsymbol{\theta}_t - \boldsymbol{\hat{\theta}}_{t|t-1})^\top {\bf P}_{t|t-1}^{-1} (\boldsymbol{\theta}_t - \boldsymbol{\hat{\theta}}_{t|t-1}),
	\end{align*}
	where 
	\begin{equation}
	\label{eq:kalman update_robust}
	\boldsymbol{\hat{\theta}}^{\text{robust}}_{t|t} = \boldsymbol{\hat{\theta}}_{t|t-1} + \mathbb{E}_{o_t \sim p(\cdot)} \big[ {\bf K}_t \big( y^{\text{robust}}_{t, \boldsymbol{\theta}'} (o_t) - Q_{}(s_t, a_t; \boldsymbol{\hat{\theta}}_{t|t-1}) \big) \big],
	\end{equation}
	and $\boldsymbol{\hat{\theta}}^{\text{robust}}_{t|t} \in \arg\min_{\boldsymbol{\theta}_t} L_t^{\text{robust EKF}} (\boldsymbol{\theta}_t)$.
\end{theorem}
The proof for Theorem \ref{theorem2} follows the same arguments as the proof for Theorem \ref{theorem1}, but uses rBTDe $\delta^{\text{robust}}_{t, \boldsymbol{\theta}'}$ (\ref{eq:robust Bellman TD error}), instead of nBTDe $\delta^{\text{nominal}}_{t, \boldsymbol{\theta}'}$. The proof appears in the supplementary material. 

Looking at the weights update in Equation (\ref{eq:kalman update_robust}) and the definition of the Kalman gain ${\bf K}_t$ in Equation (\ref{eq:Kalman_gain}), we can see that by combing the rBTDe with the EKF formulation, the Kalman gain propagates the new information from the robust target label (derived from the transition probabilities uncertainty set $\mathcal{P}$),  back down into the weight uncertainty set $\Theta$, before combining it with the estimated weight value.

This Bayesian approach for solving RMDPs is presented in Algorithm \ref{alg:Deep-RoK}, Deep-RoK. Deep-RoK receives as input the initial prior for the error covariance ${\bf P}_{0|0}$, the evolution noise covariance  ${\bf P}_{{\bf v}_t}$, the observation noise variance $P_{n_t}$, the uncertainty set $\mathcal{P}$ and a discount factor $\gamma$. Its observations are similar to the ones described for the RTD-DQN algorithm, but the weights update is different: Deep-RoK uses the EKF updates (\ref{eq:kalman update_robust}) with the robust target label $y_{j, \boldsymbol{\theta}'}^{\text{robust}}$ which is based on the uncertainty set $\mathcal{P}$.  

\begin{figure}[t]
	\centering{
		\includegraphics[width=0.45\textwidth,height=0.4\textheight,keepaspectratio]{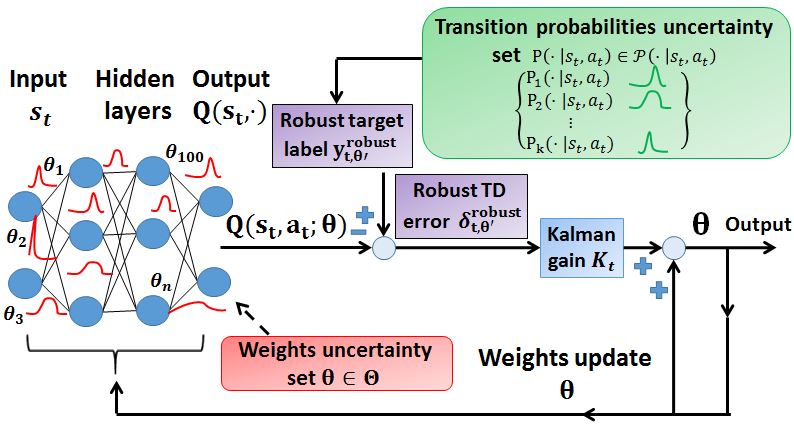}}
	\caption{Block diagram for the Deep-RoK algorithm. The Kalman gain ${\bf K}_t$ propagates the new information from the transition probabilities uncertainty set $\mathcal{P}$ into the weights uncertainty set $\Theta$, using the rBTDe $\delta^{\text{robust}}_{t, \boldsymbol{\theta}'}$.}
	\label{fig:block diagram}
\end{figure}

The output of the Deep-RoK algorithm is the MAP weights estimator $\boldsymbol{\hat{\theta}}_{t|t}^{\text{MAP}}$ and the error covariance matrix ${\bf P}_{t|t}$.  Deep-RoK is suitable for any prior ${\bf P}_{0|0}$, including priors that assume correlations between the weights. Figure \ref{fig:block diagram} presents a block diagram which illustrates the flow of weights updates in the Deep-RoK algorithm.  

During the test phase, the output of the Deep-RoK algorithm provides flexibility to the agent. It can choose to use the point estimate $\boldsymbol{\hat{\theta}}_{t|t}$ as a single NN on which it performs tests. Recall that $\boldsymbol{\hat{\theta}}_{t|t}$ incorporates the information regarding the weights uncertainty set $\Theta$ and the transitions uncertainty set $\mathcal{P}$. However, the agent has the ability to take advantage of the additional output ${\bf P}_{t|t}$ and to test the results over an ensemble of NNs by sampling weights from a distribution with mean $\boldsymbol{\hat{\theta}}_{t|t}$ and covariance ${\bf P}_{t|t}$.

\section{Experiments}
\label{Section:Experiments}
We demonstrate the performance of the RTD-DQN and Deep-RoK algorithms on the classic RL environment Cart-Pole with nonlinear Q-function approximations. In the Cart-Pole domain the agent's goal is to balance a pole atop a cart while controlling the direction of the force applied on the cart. The action space is discrete and contains two possible actions: applying a constant force to the {\it right} or to the {\it left}. The state space is continuous, where each state is four-dimensional $(x, \dot{x}, \theta, \dot{\theta})$ consisting respectively of the cart position, cart velocity, pole angle and the pole's angular velocity. At each time step the agent receives an immediate reward of 1 if the pole has not fallen down and if the cart has not run off the right or the left boundary of the screen. If these cases occur, the agent receives a reward of 0 and the episode is terminated. The transitions follow the dynamic model of the system and are based on the parameters $\psi = $ \{cart mass, pole length\}. Additional technical details regarding this experiment can be found in the supplementary material. 

In order to introduce robustness into the Cart-Pole domain, we assumed that the parameters $\psi$ are not precisely known in advance but rather they lie in a known uncertainty set $\psi \in \Psi$. This in turn introduces uncertainty in the transition probabilities defined by the set $\mathcal{P} (\psi)$. We considered the range of $0.2-1.4$ meters for the pole length and $0.1-7$ Kg for the cart mass. At each episode we uniformly sampled  5 values from each of the parameters ranges. These samples aided us in building $\mathcal{P}$. We used the implementation CartPole-v0 contained in the OpenAI gym \cite{brockman2016openai} where we added the uncertainty set parameters into the implementation. 

We trained the agent with the RTD-DQN and the Deep-RoK algorithms, both using the rBTDe. We compare their performance with a Double-DQN agent \cite{van2016deep} that uses the nBTDe. All agents were trained for $M=700$ episodes and the results are drawn from testing the trained models over $M_{\text{test}} = 500$ episodes. 

\begin{figure}[t]
	\centering{
		\includegraphics[width=0.5\textwidth,height=0.8\textheight,keepaspectratio]{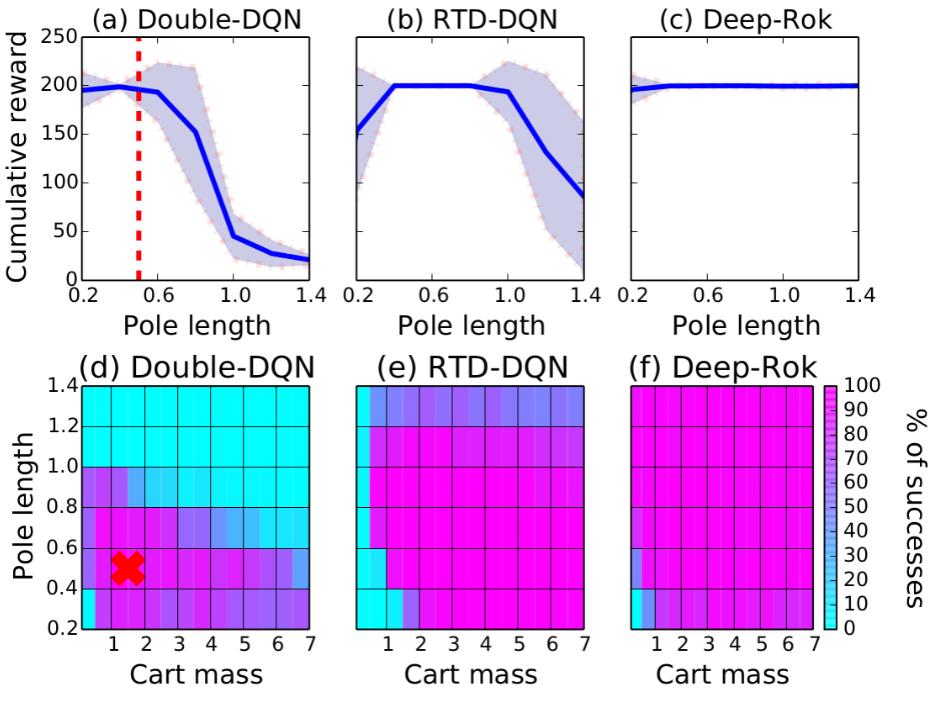}}
	\caption{{\bf (a) - (c)} Cumulative reward for different values of pole length (Mean and one standard deviation).  {\bf (d) - (f)} \% of success for different value of pole length and cart mass. A successful episode is defined by cumulative reward $> 195$. The nominal values for the Double-DQN agent  are marked in red.}
	\label{fig:comparison}
\end{figure}

In Figure \ref{fig:comparison}(a)-(c) we show the performance of all three agents for different values of pole length. In Figure \ref{fig:comparison}(d)-(f) we show the \% of success of all three agents for different values of pole length and cart mass. We can see that the Double-DQN agent performs well (high cumulative reward) on the parameters it was trained on (pole length $0.5$ meter and cart mass $1.5$ Kg, marked in red in the graphs). However it performs poorly on more extreme values of these parameters. The Double-DQN agent learned a policy which is highly optimized for the specific parameters is was trained on, but is brittle under parameter mismatch. The RTD-DQN agent has high performance also on parameter values which are not the nominal values, but they are taken into account in the uncertainty set $\mathcal{P}$ through the rBTDe. The Deep-RoK agent has the most robust results and it keeps high performance for a large range of pole lengths and cart masses. The Bayesian approach in this algorithm provides the agent with more robustness to uncertain parameters. 

\section{Discussion}
We introduced two algorithms for solving large scale RMDPs using on-line nonlinear estimations. The RTD-DQN algorithm incorporates the robust Bellman temporal difference error into a robust loss function, yielding robust policies for the agent. The Deep-RoK algorithm uses a robust Bayesian approach, based on the Extended Kalman Filter, to account for both the uncertainty in the weights of the value function and the uncertainty in the transition probabilities. We proved that the Deep-RoK algorithm outputs the weights which minimize the robust EKF lost function. We demonstrated the performance of our algorithms on the Cart-Pole domain and showed that our robust approach performs better comparing to a nominal DQN agent. We believe that real-world domains, such as autonomous driving or investment strategies, can benefit from using a robust approach to improve their agents performances by accounting for uncertainties in their models. 
 
Future work should address accounting for changes in the confidence level during the evaluation procedure, directed exploration by using uncertainty estimates and robustness in policy gradient algorithms.

\bibliography{DeepRobustKalman_arxiv}
\bibliographystyle{icml2016}

\appendix
\newpage
.
\newpage
\section*{Supplementary Material}
\section{Extended Kalman Filter (EKF)}
\label{suppSection:EKF}
In this section we briefly outline formulation of the Extended Kalman filter \cite{anderson1979optimal, gelb1974applied}. The EKF considers the following model:
\begin{equation}
\label{suppeq:Extended-Kalman}
\begin{cases}
\boldsymbol{\theta}_t = \boldsymbol{\theta}_{t-1} + {\bf v}_t\\
o_t = h(\boldsymbol{\theta}_t) + n_t
\end{cases},
\end{equation}
where $\boldsymbol{\theta}_t$ are the weights evaluated at time $t$, $o_t$ is the observation at time $t$ and $h(\boldsymbol{\theta}_t)$ is a nonlinear observation function. 

The evolution noise ${\bf v}_t$ is white ($\mathbb{E}[{\bf v}_t] = {\bf 0}$) with covariance ${\bf P}_{{\bf v}_t} \triangleq \mathbb{E}[{\bf v}_t {\bf v}_t^\top]$, $\ \mathbb{E}[{\bf v}_t {\bf v}_{t'}^\top] = {\bf 0}, \quad \forall t \neq t'$.

The observation noise $n_t$ is white ($\mathbb{E}[n_{t}]=0$) with variance $P_{n_t} \triangleq \mathbb{E}[n_t^2] $, $\ \mathbb{E}[n_{t}n_{t'}]=0, \forall t \ne t'$. 

The EKF sets the estimation of the weights $\boldsymbol{\theta}$ at time $t$ according to the conditional expectation:
\begin{align}
\label{suppeq:weight_estimation}
\nonumber \boldsymbol{\hat{\theta}}_{t|t}  & \triangleq \mathbb{E} [ \boldsymbol{\theta}_t | o_{1:t}]\\
\boldsymbol{\hat{\theta}}_{t|t-1}  & \triangleq \mathbb{E} [ \boldsymbol{\theta}_t | o_{1:t-1}]  = \boldsymbol{\hat{\theta}}_{t-1|t-1} 
\end{align}

where $o_{1:t'}$ are the observations gathered up to time $t'$. The {\it weights errors} are defined by: 
\[\boldsymbol{\tilde{\theta}}_{t|t}  \triangleq \boldsymbol{\theta}_t - \boldsymbol{\hat{\theta}}_{t|t} \]
\begin{equation}
\label{suppeq:weights_error}
\boldsymbol{\tilde{\theta}}_{t|t-1} \triangleq \boldsymbol{\theta}_t - \boldsymbol{\hat{\theta}}_{t|t-1}
\end{equation}

The conditional {\it error covariances} are given by: 
\begin{align}
{\bf P}_{t|t}  & \triangleq \mathbb{E} \big[  \boldsymbol{\tilde{\theta}}_{t|t}  \boldsymbol{\tilde{\theta}}_{t|t}^\top | o_{1:t}\big], \label{suppeq:error_covariance_prediction} \\
{\bf P}_{t|t-1} & \triangleq \mathbb{E} \big[  \boldsymbol{\tilde{\theta}}_{t|t-1}  \boldsymbol{\tilde{\theta}}_{t|t-1}^\top | o_{1:t-1}\big] \label{suppeq:error_covariance} \\
\nonumber & = \mathbb{E} \big[  (\boldsymbol{\theta}_t - \boldsymbol{\hat{\theta}}_{t|t-1}) (\boldsymbol{\theta}_t - \boldsymbol{\hat{\theta}}_{t|t-1})^\top | o_{1:t-1}\big]\\
\nonumber & = \mathbb{E} \big[  (\boldsymbol{\theta}_{t-1} + {\bf v}_t - \boldsymbol{\hat{\theta}}_{t-1|t-1}) \\
\nonumber & \quad \quad \quad (\boldsymbol{\theta}_{t-1} + {\bf v}_t - \boldsymbol{\hat{\theta}}_{t-1|t-1})^\top | o_{1:t-1}\big]\\
\nonumber & \underbrace{=}_{(\ref{suppeq:weights_error})} \mathbb{E} \big[  (\boldsymbol{\tilde{\theta}}_{t-1|t-1} + {\bf v}_t) (\boldsymbol{\tilde{\theta}}_{t-1|t-1} + {\bf v}_t)^\top | o_{1:t-1}\big]\\
\nonumber & = \mathbb{E} \big[  (\boldsymbol{\tilde{\theta}}_{t-1|t-1} \boldsymbol{\tilde{\theta}}_{t-1|t-1}^\top | o_{1:t-1}\big]\\
\nonumber & \quad +2 \mathbb{E} \big[ \boldsymbol{\tilde{\theta}}_{t-1|t-1} {\bf v}_t^\top| o_{1:t-1} \big] + \mathbb{E} \big[ {\bf v}_t {\bf v}_t^\top | o_{1:t-1}\big]\\
\nonumber & \underbrace{=}_{(\ref{suppeq:error_covariance_prediction})} {\bf P}_{t-1|t-1} + {\bf P}_{{\bf v}_t}.
\end{align}
\begin{equation*}
\boxed{{\bf P}_{t|t-1} = {\bf P}_{t-1|t-1} + {\bf P}_{{\bf v}_t}}
\end{equation*}
EKF considers several statistics of interest at each time step:
{\it The prediction of the observation function}:
\[\hat{o}_{t|t-1} \triangleq  \mathbb{E}[h(\boldsymbol{\theta}_t)|o_{1:t-1}].\]
{\it The observation innovation}:
\[\tilde{o}_{t|t-1} \triangleq h(\boldsymbol{\theta}_t) - \hat{o}_{t|t-1}.\]
{\it The covariance between the weights error and the innovation}:
\[{\bf P}_{\boldsymbol{\tilde{\theta}}_t,\tilde{o}_t}  \triangleq \mathbb{E}[ \boldsymbol{\tilde{\theta}}_{t|t-1} \tilde{o}_{t|t-1} |o_{1:t-1}].\]
{\it The covariance of the innovation}:
\[P_{\tilde{o}_t}  \triangleq \mathbb{E}[( \tilde{o}_{t|t-1})^2|o_{1:t-1}] + P_{n_t}. \]
The {\it Kalman gain}:
\[{\bf K}_t \triangleq {\bf P}_{\boldsymbol{\tilde{\theta}}_t,\tilde{o}_t} P_{\tilde{o}_t}^{-1}.\]
The above statistics serve for the update of the weights and the error covariance:
\[\begin{cases}
\boldsymbol{\hat{\theta}}_{t|t} = \boldsymbol{\hat{\theta}}_{t|t-1} +  \mathbb{E}_{o_t \sim p(\cdot)} \big[ {\bf K}_t \big( o_t - h(\boldsymbol{\hat{\theta}}_{t|t-1}) \big) \big],\\
{\bf P}_{t|t} = {\bf P}_{t|t-1} - {\bf K}_t P_{\tilde{o}_t} {\bf K}_t^\top.
\end{cases}\]

\section{EKF for Deep Q-learning}
\label{suppSection:EKF_Qlearning}
When applying the EKF formulation to Deep Q-leaning networks, the observation at time $t$ is simply the nominal target label $y^{\text{nominal}}_{t, \boldsymbol{\theta}'} (o_t)$:
\begin{equation}
\label{suppeq:target label DQN}
y^{\text{nominal}}_{t, \boldsymbol{\theta}'} (o_t) \triangleq  r_t + \gamma \max_{a'} Q(s_{t+1}, a'; \boldsymbol{\theta}').
\end{equation}
where the weight $\boldsymbol{\theta}'$ is a fixed set of weights.
The observation function is the state-action Q-function:
\begin{equation}
\label{suppeq:observation_function}
h(\boldsymbol{\theta}_t) = Q(s_t, a_t; \boldsymbol{\theta}_t).
\end{equation}

We use the notation $y_t(o_t)$ for a general target label (for example $y_{t, \boldsymbol{\theta}'}^{\text{nominal}}(o_t)$ or $y_{t, \boldsymbol{\theta}'}^{\text{robust}}(o_t)$) that serves as an observation in the EKF formulation. With this formulation, the EKF model for DQN with Bayesian approach becomes:
\begin{equation}
\label{suppeq:Extended Kalman_DQN}
\begin{cases}
\boldsymbol{\theta}_t = \boldsymbol{\theta}_{t-1} + {\bf v}_t\\
y_{t} (o_t) = Q(s_t, a_t; \boldsymbol{\theta}_t) + n_t
\end{cases}.
\end{equation}
The EKF uses a first order Taylor series linearization for the observation function (Q-function):\\
\begin{equation}
\label{suppeq:Linearization}
Q(s_t, a_t; \boldsymbol{\theta}_t) 
= Q(s_t, a_t; \boldsymbol{\hat{\theta}}) +  \big( \boldsymbol{\theta}_{t} - \boldsymbol{\hat{\theta}} \big)^\top \nabla_{\boldsymbol{\theta}_t} Q(s_t, a_t; \boldsymbol{\hat{\theta}}),
\end{equation}
where $\boldsymbol{\hat{\theta}}$ is typically chosen to be the previous estimation of the weights at time $t-1$,  $\boldsymbol{\hat{\theta}}_{t|t-1}$. This linearization helps in computing the statistics of interest. Recall that the expectation here is over the random variable $\boldsymbol{\theta}_t$ where $\boldsymbol{\hat{\theta}}_{t|t-1}$ is fixed. 
{\it The prediction of the observation function}:
\begin{align*}
\hat{y}_{t|t-1}(o_t) & \triangleq  \mathbb{E}[h(\boldsymbol{\theta}_t)|o_{1:t-1}]\\
& \underbrace{=}_{(\ref{suppeq:observation_function})} \mathbb{E}[Q(s_t, a_t; \boldsymbol{\theta}_t)|o_{1:t-1}]\\
& \underbrace{=}_{(\ref{suppeq:Linearization})} \mathbb{E}[Q(s_t, a_t; \boldsymbol{\hat{\theta}}_{t|t-1}) \\
& \quad +  \big( \boldsymbol{\theta}_{t} - \boldsymbol{\hat{\theta}}_{t|t-1} \big)^\top \nabla_{\boldsymbol{\theta}_t} Q(s_t, a_t; \boldsymbol{\hat{\theta}}_{t|t-1})|o_{1:t-1}]\\
& = Q(s_t, a_t; \boldsymbol{\hat{\theta}}_{t|t-1})  +  \big( \mathbb{E}[  \boldsymbol{\theta}_{t} |o_{1:t-1}] - \boldsymbol{\hat{\theta}}_{t|t-1} \big)^\top \\ & \quad \nabla_{\boldsymbol{\theta}_t} Q(s_t, a_t; \boldsymbol{\hat{\theta}}_{t|t-1})\\
& \underbrace{=}_{(\ref{suppeq:weight_estimation})} Q(s_t, a_t; \boldsymbol{\hat{\theta}}_{t|t-1})  +  \big( \boldsymbol{\hat{\theta}}_{t|t-1} - \boldsymbol{\hat{\theta}}_{t|t-1} \big)^\top \\
& \quad \nabla_{\boldsymbol{\theta}_t} Q(s_t, a_t; \boldsymbol{\hat{\theta}}_{t|t-1})|o_{1:t-1}]\\
& = Q(s_t, a_t; \boldsymbol{\hat{\theta}}_{t|t-1})
\end{align*}
We conclude that:
\begin{equation}
\label{suppeq:prediction_observation}
\boxed{ \hat{y}_{t|t-1}(o_t)   = Q(s_t, a_t; \boldsymbol{\hat{\theta}}_{t|t-1}) }
\end{equation}
{\it The observation innovation}:
\begin{align*}
\tilde{y}_{t|t-1}(o_t) & \triangleq h(\boldsymbol{\theta}_t) - \hat{y}_{t|t-1}(o_t) \\
& \underbrace{=}_{(\ref{suppeq:observation_function}) + (\ref{suppeq:prediction_observation})} Q(s_t, a_t; \boldsymbol{\theta}_t) - Q(s_t, a_t; \boldsymbol{\hat{\theta}}_{t|t-1}) 
\end{align*}
\begin{equation}
\label{suppeq:observation_innovation}
\boxed{\tilde{y}_{t|t-1}(o_t) = Q(s_t, a_t; \boldsymbol{\theta}_t) - Q(s_t, a_t; \boldsymbol{\hat{\theta}}_{t|t-1}) }
\end{equation}
{\it The covariance between the weights error and the innovation}:
\begin{align*}
& {\bf P}_{\boldsymbol{\tilde{\theta}}_t,\tilde{y}(o_t)} \triangleq \mathbb{E}[ \boldsymbol{\tilde{\theta}}_{t|t-1} \tilde{y}_{t|t-1}(o_t)  |o_{1:t-1}]\\
& \underbrace{=}_{(\ref{suppeq:weights_error}) + (\ref{suppeq:observation_innovation})} \mathbb{E}[ \big(\boldsymbol{\theta}_t - \boldsymbol{\hat{\theta}}_{t|t-1} \big)\\
& \quad \quad \big( Q(s_t, a_t; \boldsymbol{\theta}_t) - Q(s_t, a_t; \boldsymbol{\hat{\theta}}_{t|t-1}) \big) |o_{1:t-1}]\\
& \underbrace{=}_{(\ref{suppeq:Linearization})} \mathbb{E}[ \big(\boldsymbol{\theta}_t - \boldsymbol{\hat{\theta}}_{t|t-1} \big) \big( \cancel{Q(s_t, a_t; \boldsymbol{\hat{\theta}}_{t|t-1})}  +  \big( \boldsymbol{\theta}_{t} - \boldsymbol{\hat{\theta}}_{t|t-1} \big)^\top \\
& \quad  \nabla_{\boldsymbol{\theta}_t} Q(s_t, a_t; \boldsymbol{\hat{\theta}}_{t|t-1}) - \cancel{Q(s_t, a_t; \boldsymbol{\hat{\theta}}_{t|t-1}) \big)} |o_{1:t-1}]\\
& \underbrace{=}_{(\ref{suppeq:weights_error})} \mathbb{E}[ \boldsymbol{\tilde{\theta}}_{t|t-1}  \boldsymbol{\tilde{\theta}}_{t|t-1}^\top  |o_{1:t-1}] \nabla_{\boldsymbol{\theta}_t} Q(s_t, a_t; \boldsymbol{\hat{\theta}}_{t|t-1})\\
& \underbrace{=}_{(\ref{suppeq:error_covariance})} {\bf P}_{t|t-1} \nabla_{\boldsymbol{\theta}_t} Q(s_t, a_t; \boldsymbol{\hat{\theta}}_{t|t-1})
\end{align*}
\begin{equation}
\label{suppeq:covariance_weights_innovation}
\boxed{{\bf P}_{\boldsymbol{\tilde{\theta}}_t,\tilde{y}(o_t)} = {\bf P}_{t|t-1} \nabla_{\boldsymbol{\theta}_t} Q(s_t, a_t; \boldsymbol{\hat{\theta}}_{t|t-1})}
\end{equation}
{\it The covariance of the innovation}:
\begin{align*}
& P_{\tilde{y}(o_t)}  \triangleq \mathbb{E}[ \big( \tilde{y}_{t|t-1}(o_t) \big)^2|o_{1:t-1}] + P_{n_t}\\
& \underbrace{=}_{(\ref{suppeq:observation_innovation})} \mathbb{E}[ \big( Q(s_t, a_t; \boldsymbol{\theta}_t) - Q(s_t, a_t; \boldsymbol{\hat{\theta}}_{t|t-1}) \big)^2|o_{1:t-1}] + P_{n_t}\\
& \underbrace{=}_{(\ref{suppeq:Linearization})} \mathbb{E}[ \big(\cancel{Q(s_t, a_t; \boldsymbol{\hat{\theta}}_{t|t-1})}  +  \big( \boldsymbol{\theta}_{t} - \boldsymbol{\hat{\theta}}_{t|t-1} \big)^\top \\
& \quad  \nabla_{\boldsymbol{\theta}_t} Q(s_t, a_t; \boldsymbol{\hat{\theta}}_{t|t-1}) - \cancel{Q(s_t, a_t; \boldsymbol{\hat{\theta}}_{t|t-1})} \big)^2|o_{1:t-1}] + P_{n_t}\\
& = \mathbb{E}[ \big(\big( \boldsymbol{\theta}_{t} - \boldsymbol{\hat{\theta}}_{t|t-1} \big)^\top   \nabla_{\boldsymbol{\theta}_t} Q(s_t, a_t; \boldsymbol{\hat{\theta}}_{t|t-1}) \big)^2|o_{1:t-1}] + P_{n_t}\\
& \underbrace{=}_{(\ref{suppeq:weights_error})} \nabla_{\boldsymbol{\theta}_t} Q(s_t, a_t; \boldsymbol{\hat{\theta}}_{t|t-1})^\top \mathbb{E}[ \boldsymbol{\tilde{\theta}}_{t|t-1}  \boldsymbol{\tilde{\theta}}_{t|t-1}^\top    |o_{1:t-1}] \\
&\quad \quad \nabla_{\boldsymbol{\theta}_t} Q(s_t, a_t; \boldsymbol{\hat{\theta}}_{t|t-1}) + P_{n_t}\\
& \underbrace{=}_{(\ref{suppeq:error_covariance})} \nabla_{\boldsymbol{\theta}_t} Q(s_t, a_t; \boldsymbol{\hat{\theta}}_{t|t-1})^\top {\bf P}_{t|t-1} \nabla_{\boldsymbol{\theta}_t} Q(s_t, a_t; \boldsymbol{\hat{\theta}}_{t|t-1}) + P_{n_t}
\end{align*}
\begin{empheq}[box=\fbox]{align}
\label{suppeq:covariance_innovation}
P_{\tilde{y}(o_t)} &= \nabla_{\boldsymbol{\theta}_t} Q(s_t, a_t; \boldsymbol{\hat{\theta}}_{t|t-1})^\top {\bf P}_{t|t-1}  \nonumber\\
& \quad \quad \quad \quad \quad \quad \quad  \quad  \nabla_{\boldsymbol{\theta}_t} Q(s_t, a_t; \boldsymbol{\hat{\theta}}_{t|t-1}) + P_{n_t}
\end{empheq}
The {\it Kalman gain}:
\begin{align}
\label{suppeq:kalman_gain}
\nonumber & {\bf K}_t  \triangleq {\bf P}_{\boldsymbol{\tilde{\theta}}_t,\tilde{y}(o_t)} P_{\tilde{y}(o_t)}^{-1}\\
& \underbrace{=}_{(\ref{suppeq:covariance_weights_innovation}) + (\ref{suppeq:covariance_innovation})} {\bf P}_{t|t-1} \nabla_{\boldsymbol{\theta}_t} Q(s_t, a_t; \boldsymbol{\hat{\theta}}_{t|t-1})\\
\nonumber & \Big( \nabla_{\boldsymbol{\theta}_t} Q(s_t, a_t; \boldsymbol{\hat{\theta}}_{t|t-1})^\top {\bf P}_{t|t-1} \nabla_{\boldsymbol{\theta}_t} Q(s_t, a_t; \boldsymbol{\hat{\theta}}_{t|t-1})  + P_{n_t} \Big)^{-1}
\end{align}
The update for the weights of the Q-function and the error covariance are:
\begin{align}
\label{suppeq:kalman update}
\begin{cases}
\boldsymbol{\hat{\theta}}^{\text{EKF}}_{t|t} = \boldsymbol{\hat{\theta}}_{t|t-1} +  \mathbb{E}_{o_t \sim p(\cdot)} \big[ {\bf K}_t \big( y^{\text{nominal}}_{t, \boldsymbol{\theta}'} (o_t) \\
\quad \quad \quad \quad \quad \quad \quad \quad \quad \quad  - Q_{}(s_t, a_t; \boldsymbol{\hat{\theta}}_{t|t-1}) \big) \big]\\
{\bf P}_{t|t} = {\bf P}_{t|t-1} - {\bf K}_t P_{\tilde{y}(o_t)} {\bf K}_t^\top.
\end{cases}
\end{align}
as we prove in Theorem \ref{theorem1}. 
\section{A Bayesian approach: MAP estimator}
We adopt the Bayesian approach in which we are interested in finding the optimal set of weights $\boldsymbol{\theta}_t$ that maximizes the posterior distribution of the weights given the observations we have gathered up to time $t$, denoted as the  $o_{1:t}$.

According to Bayes rule, the posterior distribution is defined as: 
\[p(\boldsymbol{\theta}_t|o_{1:t}) = \frac{p(o_{1:t}|\boldsymbol{\theta}_t) p(\boldsymbol{\theta}_t)}{p(o_{1:t})}\]
where $p(o_{1:t}|\boldsymbol{\theta})$ is the {\it likelihood} of the observations given the weights $\boldsymbol{\theta}$ and $p(\boldsymbol{\theta})$ is the {\it prior} distribution over $\boldsymbol{\theta}$.	We will expend the term of the posterior \cite{van2004sigma}:
\begin{align}
\nonumber p(\boldsymbol{\theta}_t|o_{1:t}) & = \frac{p(o_{1:t}|\boldsymbol{\theta}_t)p(\boldsymbol{\theta}_t)}{p(o_{1:t})} \\
& = \frac{p(o_t|o_{1:t-1},\boldsymbol{\theta}_t) p(o_{1:t-1}|\boldsymbol{\theta}_t)p(\boldsymbol{\theta}_t)}{p(o_{1:t})} \label{eq:PosteriorWithNoise1} \\ 
& = \frac{p(o_t|\boldsymbol{\theta}_t) p(o_{1:t-1}|\boldsymbol{\theta}_t)p(\boldsymbol{\theta}_t)  }{p(o_{1:t})} \cdot \frac{p(o_{1:t-1})}{p(o_{1:t-1})}  \label{eq:PosteriorWithNoise2}\\ 
& = \frac{p(o_t|\boldsymbol{\theta}_t)p(\boldsymbol{\theta}_t|o_{1:t-1}) p(o_{1:t-1}) }{p(o_{1:t})} \label{eq:PosteriorWithNoise3}
\end{align}
The transition in (\ref{eq:PosteriorWithNoise1}) is according to the conditional probability:
\begin{align*}
p(o_{1:t}|\boldsymbol{\theta}_t) & = p(o_t, o_{1:t-1}|\boldsymbol{\theta}_t)\\
& = \frac{p(o_t, o_{1:t-1},\boldsymbol{\theta}_t)}{p(\boldsymbol{\theta}_t)} \\
& = \frac{p(o_{1:t-1},\boldsymbol{\theta}_t)p(o_t|o_{1:t-1},\boldsymbol{\theta}_t)}{p(\boldsymbol{\theta}_t)} \\
& = p(o_{1:t-1}|\boldsymbol{\theta}_t)p(o_t|o_{1:t-1},\boldsymbol{\theta}_t) 
\end{align*}
The transition in (\ref{eq:PosteriorWithNoise2}) is according to the conditional independence: $p(o_t|o_{1:t-1},\boldsymbol{\theta}_t) = p(o_t|\boldsymbol{\theta}_t)$, and we multiplied the numerator and the dominator by $p(o_{1:t-1})$.\\
The transition in (\ref{eq:PosteriorWithNoise3}) is according to Bayes rule: $p(\boldsymbol{\theta}_t|o_{1:t-1}) = \frac{p(o_{1:t-1}|\boldsymbol{\theta}_t)p(\boldsymbol{\theta}_t)}{p(o_{1:t-1})}$.

The MAP estimator for $\boldsymbol{\theta}_t$ is the one who maximizes the posterior distribution described in (\ref{eq:PosteriorWithNoise3}).
\begin{align}
\nonumber \boldsymbol{\theta}_{t}^{MAP}  = & \arg\max_{\boldsymbol{\theta}_t} && \big\{ p(\boldsymbol{\theta}_t|o_{1:t}) \big\}\\
\nonumber = &\arg\max_{\boldsymbol{\theta}_t} && \Big\{  \frac{p(o_t|\boldsymbol{\theta}_t)p(\boldsymbol{\theta}_t|o_{1:t-1}) p(o_{1:t-1}) }{p(o_{1:t})} \Big\}\\
\nonumber = &\arg\max_{\boldsymbol{\theta}_t} && \big\{  p(o_t|\boldsymbol{\theta}_t)p(\boldsymbol{\theta}_t|o_{1:t-1})  \big\}\\
\nonumber = &\arg\max_{\boldsymbol{\theta}_t} && \big\{ \log \big(  p(o_t|\boldsymbol{\theta}_t)p(\boldsymbol{\theta}_t|o_{1:t-1})  \big) \big\} \\
\nonumber = &\arg\max_{\boldsymbol{\theta}_t} && \big\{ \log   p(o_t|\boldsymbol{\theta}_t) + \log p(\boldsymbol{\theta}_t|o_{1:t-1})   \big\} \\
= &\arg\min_{\boldsymbol{\theta}_t} && \big\{ - \log   p(o_t|\boldsymbol{\theta}_t) - \log p(\boldsymbol{\theta}_t|o_{1:t-1})   \big\} \label{eq:MAPln}
\end{align}
In (\ref{eq:MAPln}) We used the derivation in ($\ref{eq:PosteriorWithNoise3}$) and the fact that the argument which maximizes the posterior is the same as the argument that maximizes the $\log (\cdot)$ of the posterior. In addition this argument also minimizes the negative $\log(\cdot)$. 

We will replace here the current observation $o_t$ with the current target label $y_t(o_t)$ and receive:
\begin{equation}
\label{suppeq:MAP}
\boldsymbol{\theta}_{t}^{MAP} = \arg\min_{\boldsymbol{\theta}_t}  \big\{ - \log   p(y_t(o_t)|\boldsymbol{\theta}_t) - \log p(\boldsymbol{\theta}_t|o_{1:t-1})   \big\}
\end{equation}

\section{Gaussian assumptions}
When estimating using the EKF, it is common to make the following assumptions regarding the likelihood and the posterior in Equation (\ref{suppeq:MAP}):\\
{\bf Assumption 1}. \textit{The likelihood $p(y_t(o_t)|\boldsymbol{\theta}_t)$ is assumed to be Gaussian: 
	$y_t(o_t)|\boldsymbol{\theta}_t \sim \mathcal{N}(\mathbb{E}_{o_t \sim p(\cdot)} [Q(s_t, a_t; \boldsymbol{\theta}_t)], P_{n_t})$.}\\
{\bf Assumption 2}. \textit{The posterior distribution $p(\boldsymbol{\theta}_t|o_{1:t-1})$ is assumed to be Gaussian: $\boldsymbol{\theta}_t|o_{1:t-1} \sim \mathcal{N}(\boldsymbol{\hat{\theta}}_{t|t-1},{\bf P}_{t|t-1})$.}

Following are the calculations for the means and covariances in Assumptions \ref{As:ConditionalIndependance} and \ref{As:GaussianPosterior}. For the likelihood $p(y_t(o_t)|\boldsymbol{\theta}_t)$:
\begin{align*}
\mathbb{E}_{o_t} \big[y_t(o_t)|\boldsymbol{\theta}_t \big] & \underbrace{=}_{(\ref{suppeq:Extended Kalman_DQN})} \mathbb{E}_{o_t} \big[Q(s_t, a_t; \boldsymbol{\theta}_t) + n_t|\boldsymbol{\theta}_t \big]\\
&  = \mathbb{E}_{o_t} \big[Q(s_t, a_t; \boldsymbol{\theta}_t) |\boldsymbol{\theta}_t \big] + \underbrace{\mathbb{E}_{o_t} \big[ n_t |\boldsymbol{\theta}_t\big]}_{=0}\\
& = \mathbb{E}_{o_t} \big[Q(s_t, a_t; \boldsymbol{\theta}_t) \big]
\end{align*}
\begin{align*}
& Cov(y_t(o_t)|\boldsymbol{\theta}_t)  \triangleq 
\mathbb{E}_{o_t} \big[\big(y(o_t) - \mathbb{E}_{o_t} \big[y(o_t)|\boldsymbol{\theta}_t \big] \big)^2|\boldsymbol{\theta}_t \big] \\
& \underbrace{=}_{(\ref{suppeq:Extended Kalman_DQN})} \mathbb{E}_{o_t} \big[\big(Q(s_t, a_t; \boldsymbol{\theta}_t) + n_t - \mathbb{E}_{o_t} \big[Q(s_t, a_t; \boldsymbol{\theta}_t) \big] \big)^2|\boldsymbol{\theta}_t \big]\\
& = \mathbb{E} \big[ n_t^2]\\
& = P_{n_t}
\end{align*}
For the posterior $p(\boldsymbol{\theta}_t|o_{1:t-1})$:
\begin{equation*}
\mathbb{E}_{\boldsymbol{\theta}_t} \big[ \boldsymbol{\theta}_t|o_{1:t-1}\big] \underbrace{=}_{(\ref{suppeq:weight_estimation})} \boldsymbol{\hat{\theta}}_{t|t-1}
\end{equation*} 
\begin{align*}
Cov(\boldsymbol{\theta}_t|o_{1:t-1}) \triangleq  
& \mathbb{E}_{\boldsymbol{\theta}_t} \big[ \big(\boldsymbol{\theta}_t - \boldsymbol{\hat{\theta}}_{t|t-1} \big) \big(\boldsymbol{\theta}_t - \boldsymbol{\hat{\theta}}_{t|t-1} \big)^\top |o_{1:t-1} \big]\\
& \underbrace{=}_{(\ref{suppeq:weights_error})} \mathbb{E}_{\boldsymbol{\theta}_t} \big[ \boldsymbol{\tilde{\theta}}_{t|t-1} \boldsymbol{\tilde{\theta}}_{t|t-1}^\top |o_{1:t-1} \big] \underbrace{=}_{(\ref{suppeq:error_covariance})} {\bf P}_{t|t-1}
\end{align*}

\section{Theoretical results}
Based on the Gaussian assumptions, we can derive the following Theorem:\\
{\bf Theorem 1.} \textit{Under Assumptions \ref{As:ConditionalIndependance} and \ref{As:GaussianPosterior}, $\boldsymbol{\hat{\theta}}^{\text{EKF}}_{t|t}$  (\ref{suppeq:kalman update}) minimizes at each time step $t$ the following regularized objective function:
	\begin{align}
	\label{suppeq:objective_EKF}
	\nonumber L^{\text{EKF}}_t(\boldsymbol{\theta}_t) & =  \frac{1}{2P_{n_t}} \mathbb{E}_{o_t \sim p(\cdot)} [ \big(\delta^{\text{nominal}}_{t, \boldsymbol{\theta}'} (\boldsymbol{\theta}_t, o_t) \big)^2] \\
	& +  \frac{1}{2}(\boldsymbol{\theta}_t - \boldsymbol{\hat{\theta}}_{t|t-1})^\top {\bf P}_{t|t-1}^{-1} (\boldsymbol{\theta}_t - \boldsymbol{\hat{\theta}}_{t|t-1}),
	\end{align}
	where $\boldsymbol{\hat{\theta}}^{\text{EKF}}_{t|t} \in \arg\min_{\boldsymbol{\theta}_t} L^{\text{EKF}}_t(\boldsymbol{\theta}_t)$.}

\begin{proof}
	We solve the minimization problem in (\ref{suppeq:MAP}) by substituting the Gaussian Assumptions \ref{As:ConditionalIndependance} and \ref{As:GaussianPosterior}. We show that this minimization problem is equivalent to minimize the objective function $L_t^{\text{EKF}}$ in Theorem \ref{theorem1}. 
	\small
	\begin{align*}
	\nonumber \boldsymbol{\hat{\theta}}_{t|t}^{\text{MAP}} & = \arg\min_{\boldsymbol{\theta}_t}  \big\{  - \log \Big( p(y_t(o_t)|\boldsymbol{\theta}_t) \Big) - \log \Big( p(\boldsymbol{\theta}_t|o_{1:t-1}) \Big) \big\}\\
	\nonumber & = \arg\min_{\boldsymbol{\theta}_t}  \Big\{ -\log \bigg( \frac{1}{\sqrt{2\pi P_{n_t}} } \\
	\nonumber & \quad \exp \Big( - \frac{1}{2P_{n_t}} \big(y_t(o_t) - \mathbb{E}_{o_t} \big[Q(s_t, a_t; \boldsymbol{\theta}_t) \big] \big)^2   \Big) \bigg) \\
	\nonumber& -\log \bigg( \frac{1}{(2 \pi)^{n/2} |{\bf P}_{t|t-1}|^{1/2}}  \\
	\nonumber & \quad \exp \Big( - \frac{1}{2} (\boldsymbol{\theta}_t - \boldsymbol{\hat{\theta}}_{t|t-1})^\top  {\bf P}_{t|t-1}^{-1} (\boldsymbol{\theta}_t -  \boldsymbol{\hat{\theta}}_{t|t-1})  \Big) \bigg)  \\
	\nonumber & = \arg\min_{\boldsymbol{\theta}_i}  \Big\{ \frac{1}{2P_{n_t}} \Big(y_t(o_t) - \mathbb{E}_{o_t \sim p(\cdot)} [Q(s_t, a_t; \boldsymbol{\theta}_t)]  \Big)^2 \\
	\nonumber & \quad  -\log \Big( \frac{1}{\sqrt{2\pi P_{n_t}} } \Big)  \\
	\nonumber & + \frac{1}{2} (\boldsymbol{\theta}_t - \boldsymbol{\hat{\theta}}_{t|t-1})^\top  {\bf P}_{t|t-1}^{-1} (\boldsymbol{\theta}_t -  \boldsymbol{\hat{\theta}}_{t|t-1})\\
	\nonumber & \quad  -\log \Big( \frac{1}{(2 \pi)^{n/2} |{\bf P}_{t|t-1}|^{1/2}}  \Big) \Big\} \\
	\nonumber & = \arg\min_{\boldsymbol{\theta}_i}  \Big\{ \frac{1}{2P_{n_t}} \Big(y_t(o_t) - \mathbb{E}_{o_t \sim p(\cdot)} [Q(s_t, a_t; \boldsymbol{\theta}_t)]  \Big)^2 \\
	\nonumber & + \frac{1}{2} (\boldsymbol{\theta}_t - \boldsymbol{\hat{\theta}}_{t|t-1})^\top  {\bf P}_{t|t-1}^{-1} (\boldsymbol{\theta}_t -  \boldsymbol{\hat{\theta}}_{t|t-1}) \Big\} 
	\end{align*}
	\normalsize
	where $|\cdot|$ denotes the determinant. We receive the following objective function:
	\begin{align}
	\label{eq:Objective}
	\nonumber L_t(\boldsymbol{\theta}_t) & =  \frac{1}{2P_{n_t}} \mathbb{E}_{o_t \sim p(\cdot)} \Big[ \Big(y_t(o_t) -  Q(s_t, a_t; \boldsymbol{\theta}_t) \Big)^2 \Big] \\
	\nonumber & + \frac{1}{2} (\boldsymbol{\theta}_t - \boldsymbol{\hat{\theta}}_{t|t-1})^\top  {\bf P}_{t|t-1}^{-1} (\boldsymbol{\theta}_t -  \boldsymbol{\hat{\theta}}_{t|t-1}).  
	\end{align}
	We replace $y_t(o_t)$ with the nominal target label $y^{\text{nominal}}_{t, \boldsymbol{\theta}'} (o_t)$ and define the nominal Bellman TD error $\delta^{\text{nominal}}_{t, \boldsymbol{\theta}'}$:
	\begin{equation}
	\label{suppeq: Bellman TD}
	\delta^{\text{nominal}}_{t, \boldsymbol{\theta}'} (\boldsymbol{\theta}_t, o_t) \triangleq y^{\text{nominal}}_{t, \boldsymbol{\theta}'} (o_t) - Q(s_t, a_t; \boldsymbol{\theta}_t).
	\end{equation}
	We receive:
	\begin{align*}
	\nonumber L^{\text{EKF}}_t(\boldsymbol{\theta}_t) & =  \frac{1}{2P_{n_t}} \mathbb{E}_{o_t \sim p(\cdot)} [ \big(\delta^{\text{nominal}}_{t, \boldsymbol{\theta}'} (\boldsymbol{\theta}_t, o_t) \big)^2] \\
	& +  \frac{1}{2}(\boldsymbol{\theta}_t - \boldsymbol{\hat{\theta}}_{t|t-1})^\top {\bf P}_{t|t-1}^{-1} (\boldsymbol{\theta}_t - \boldsymbol{\hat{\theta}}_{t|t-1}),
	\end{align*}
	which is exactly the objective function (\ref{suppeq:objective_EKF}) in Theorem \ref{theorem1}. To minimize this objective function we take the derivative of $L^{\text{EKF}}_t(\boldsymbol{\theta}_t)$ with respect to $\boldsymbol{\theta}_t$:
	\small
	\begin{align*}
	\nabla_{\boldsymbol{\theta}_t} L^{\text{EKF}}_t(\boldsymbol{\theta}_t) & = - \frac{1}{P_{n_t}} \mathbb{E}_{o_t \sim p(\cdot)} [ \delta^{\text{nominal}}_{t, \boldsymbol{\theta}'} (\boldsymbol{\theta}_t, o_t) \nabla_{\boldsymbol{\theta}_t}  Q(s_t, a_t; \boldsymbol{\theta}_t) ]\\
	& \quad + {\bf P}_{t|t-1}^{-1} (\boldsymbol{\theta}_t - \boldsymbol{\hat{\theta}}_{t|t-1})  = 0
	\end{align*}
	\normalsize
	We use the linearization of the Q-function in Equation (\ref{suppeq:Linearization}) and the definition of $\delta^{\text{nominal}}_{t, \boldsymbol{\theta}'}$ in Equation (\ref{suppeq: Bellman TD}):
	\begin{align*}
	& {\bf P}_{t|t-1}^{-1} (\boldsymbol{\theta}_t - \boldsymbol{\hat{\theta}}_{t|t-1})  = \frac{1}{P_{n_t}} \mathbb{E}_{o_t \sim p(\cdot)} \bigg[ \bigg(y^{\text{nominal}}_{t, \boldsymbol{\theta}'} (o_t)\\
	& - \big( Q(s_t, a_t; \boldsymbol{\hat{\theta}}_{t|t-1})  
	+  \nabla_{\boldsymbol{\theta}_t} Q(s_t, a_t; \boldsymbol{\hat{\theta}}_{t|t-1})^\top ( \boldsymbol{\theta}_{t} - \boldsymbol{\hat{\theta}}_{t|t-1} )    \big) \bigg)\\ 
	& \nabla_{\boldsymbol{\theta}_t} \big( Q(s_t, a_t; \boldsymbol{\hat{\theta}}_{t|t-1})  
	+  \nabla_{\boldsymbol{\theta}_t} Q(s_t, a_t; \boldsymbol{\hat{\theta}}_{t|t-1})^\top ( \boldsymbol{\theta}_{t} - \boldsymbol{\hat{\theta}}_{t|t-1} )    \big) \bigg] 
	\end{align*}
	For simplicity we denote ${\bf q}$ by
	$${\bf q} = \nabla_{\boldsymbol{\theta}_t} Q(s_t, a_t; \boldsymbol{\hat{\theta}}_{t|t-1}).$$ 
	We receive that:
	\begin{align*}
	& \big({\bf P}_{t|t-1}^{-1} + \frac{1}{P_{n_t}} \mathbb{E}_{o_t \sim p(\cdot)} [{\bf q} {\bf q}^\top] \big)  \big(\boldsymbol{\theta}_t - \boldsymbol{\hat{\theta}}_{t|t-1} \big) \\
	& = \frac{1}{P_{n_t}} \mathbb{E}_{o_t \sim p(\cdot)} \big[ {\bf q} \big(y^{\text{nominal}}_{t, \boldsymbol{\theta}'} (o_t)  - Q(s_t, a_t; \boldsymbol{\hat{\theta}}_{t|t-1}) \big) \big] 
	\end{align*}
	and finally:
	\begin{align}
	\label{suppeq:weight_MAP}
	\nonumber \boldsymbol{\theta}_t & =  \boldsymbol{\hat{\theta}}_{t|t-1} +  \mathbb{E}_{o_t \sim p(\cdot)} \Big[ \frac{1}{P_{n_t}}  \big({\bf P}_{t|t-1}^{-1}  + \frac{1}{P_{n_t}}  {\bf q} {\bf q}^\top  \big)^{-1} 
	{\bf q} \\
	& \quad \big(y^{\text{nominal}}_{t, \boldsymbol{\theta}'} (o_t)  - Q(s_t, a_t; \boldsymbol{\hat{\theta}}_{t|t-1}) \big) \Big] 
	\end{align}
	We now simplify the following term:
	\small
	\begin{align}
	\label{suppeq:Kt_derivation}
	\nonumber & \frac{1}{P_{n_t}}  \big({\bf P}_{t|t-1}^{-1} + \frac{1}{P_{n_t}} {\bf q} {\bf q}^\top \big)^{-1} {\bf q} \\
	\nonumber & = \frac{1}{P_{n_t}}  \big({\bf P}_{t|t-1}^{-1} + \frac{1}{P_{n_t}} {\bf q} {\bf q}^\top \big)^{-1} {\bf q} \big( {\bf q}^\top  {\bf P}_{t|t-1}  {\bf q} + P_{n_t}\big)\\
	\nonumber & \quad \big( {\bf q}^\top  {\bf P}_{t|t-1}  {\bf q} + P_{n_t}\big)^{-1}\\
	\nonumber & = \big({\bf P}_{t|t-1}^{-1} + \frac{1}{P_{n_t}} {\bf q} {\bf q}^\top \big)^{-1} \big( \frac{1}{P_{n_t}} {\bf q} {\bf q}^\top  {\bf P}_{t|t-1}  {\bf q} + {\bf q}\big)\\
	\nonumber & \quad \big(  {\bf q}^\top  {\bf P}_{t|t-1}  {\bf q} + P_{n_t}\big)^{-1}\\
	\nonumber & = \big({\bf P}_{t|t-1}^{-1} + \frac{1}{P_{n_t}} {\bf q} {\bf q}^\top \big)^{-1} \Big( \frac{1}{P_{n_t}} {\bf q} {\bf q}^\top   + {\bf P}_{t|t-1}^{-1}\Big) {\bf P}_{t|t-1}  {\bf q}\\
	\nonumber & \quad \big(  {\bf q}^\top  {\bf P}_{t|t-1}  {\bf q} + P_{n_t}\big)^{-1}\\
	\nonumber & = {\bf P}_{t|t-1}  {\bf q} \big(  {\bf q}^\top  {\bf P}_{t|t-1}  {\bf q} + P_{n_t}\big)^{-1}\\
	\nonumber & \underbrace{=}_{(\ref{suppeq:covariance_weights_innovation}) + (\ref{suppeq:covariance_innovation})} {\bf P}_{\boldsymbol{\tilde{\theta}}_t,\tilde{y}(o_t)} P_{\tilde{y}(o_t)}^{-1}\\
	& \underbrace{=}_{(\ref{suppeq:kalman_gain})} {\bf K}_t
	\end{align}
	\normalsize
	Substituting this result in Equation (\ref{suppeq:weight_MAP}), we receive the EKF update for the weights:
	\begin{align}
	\label{suppeq:EKF_weight}
	\boldsymbol{\hat{\theta}}^{\text{EKF}}_{t|t} & = \boldsymbol{\hat{\theta}}_{t|t-1} +  \mathbb{E}_{o_t \sim p(\cdot)} \big[ {\bf K}_t \big( y^{\text{nominal}}_{t, \boldsymbol{\theta}'} (o_t) \\
	\nonumber & \quad \quad \quad \quad \quad \quad \quad \quad \quad \quad  - Q_{}(s_t, a_t; \boldsymbol{\hat{\theta}}_{t|t-1}) \big) \big]
	\end{align}
	which is exactly as in Equation (\ref{suppeq:kalman update}).
	
	We now develop the term $\big({\bf P}_{t|t-1}^{-1} + \frac{1}{P_{n_t}} {\bf q} {\bf q}^\top \big)^{-1}$ that appears in (\ref{suppeq:weight_MAP}) by using the matrix inversion lemma:
	\begin{equation}
	\label{suppeq:MatrixInversionLemma}
	({\bf B}^{-1} + {\bf C}{\bf D}^{-1}{\bf C}^\top)^{-1} = {\bf B} - {\bf BC}({\bf D}+ {\bf C}^\top {\bf BC})^{-1} {\bf C}^\top{\bf B}
	\end{equation}
	where ${\bf B}$ is a square symmetric positive-definite (and hence invertible) matrix. For this purpose we assume that the error covariance matrix of $\boldsymbol{\theta}_t$, ${\bf P}_{t|t-1}$, is symmetric and positive-definite.
	\small
	\begin{align}
	\nonumber & \Big( {\bf P}_{t|t-1}^{-1} + \frac{1}{P_{n_t}}{\bf q} {\bf q}^\top \Big)^{-1}\\
	\nonumber & \underbrace{=}_{(\ref{suppeq:MatrixInversionLemma})} {\bf P}_{t|t-1} - {\bf P}_{t|t-1} {\bf q}( P_{n_t} + {\bf q}^\top {\bf P}_{t|t-1} {\bf q})^{-1}{\bf q}^\top {\bf P}_{t|t-1} \\
	\nonumber & \underbrace{=}_{(\ref{suppeq:Kt_derivation})} {\bf P}_{t|t-1} - {\bf K}_t {\bf q}^\top {\bf P}_{t|t-1}\\
	\nonumber & \underbrace{=}_{(\ref{suppeq:covariance_weights_innovation})} {\bf P}_{t|t-1} - {\bf K}_t {\bf P}_{\boldsymbol{\tilde{\theta}}_t,\tilde{y}(o_t)}^\top\\
	\nonumber & \underbrace{=}_{(\ref{suppeq:kalman_gain})} {\bf P}_{t|t-1} - {\bf K}_t P_{\tilde{y}(o_t)} {\bf K}_t^\top
	\end{align} 
	\normalsize	
	We can write the update of the weights error covariance as:
	\begin{equation}
	\label{suppeq:error_covariance_update}
	\boxed { {\bf P}_{t|t} = {\bf P}_{t|t-1} - {\bf K}_t P_{\tilde{y}(o_t)} {\bf K}_t^\top }
	\end{equation}
	We conclude the proof by stating that the optimal weight $\boldsymbol{\hat{\theta}}_{t|t}^{\text{EKF}}$ in (\ref{suppeq:kalman update}) is the solution to the minimization of the objective function in (\ref{suppeq:objective_EKF}):
	\[\boldsymbol{\hat{\theta}}_{t|t}^{\text{EKF}} \in \arg\min_{\boldsymbol{\theta}_{t}} L_t^{\text{EKF}}(\boldsymbol{\theta}_t)\] 
\end{proof}
We are now ready to prove Theorem \ref{theorem2}. For this purpose we recall the definition of $\delta_{t, \boldsymbol{\theta}'}^{\text{robust}}$, the {\it robust Bellman TD error} and $y_{t, \boldsymbol{\theta}'}^{\text{robust}}$ the {\it robust target label}: 
\begin{equation}
\label{suppeq:robust Bellman TD error}
\delta_{t, \boldsymbol{\theta}'}^{\text{robust}}( \boldsymbol{\theta}_t, o_t) = y_{t, \boldsymbol{\theta}'}^{\text{robust}} (o_t) - Q (s_t, a_t; \boldsymbol{\theta}_t).
\end{equation}
\small
\begin{equation}
\label{suppeq:target_robust_label}
y_{t, \boldsymbol{\theta}'}^{\text{robust}}(o_t) = r_t + \gamma \min_{p \in \mathcal{P}} \sum_{s'\in \tilde{\mathcal{S}}} p(s'|s_t, a_t) \max_{a'} Q(s', a'; \boldsymbol{\theta}').
\end{equation}
\normalsize
The set $\tilde{\mathcal{S}}$ is the set of all possible next states from state $s_t$ when taking action $a_t$, and it is drawn from the uncertainty set  $\mathcal{P}_{(\cdot| s_t, a_t)}$.\\
{\bf Theorem 2.} \textit{Under Assumptions \ref{As:ConditionalIndependance} and \ref{As:GaussianPosterior}, $\boldsymbol{\hat{\theta}}^{\text{robust}}_{t|t}$  (\ref{eq:kalman update_robust}) minimizes at each time step $t$ the following regularized {\it robust objective function}:
	\begin{align*}
	\nonumber L_t^{\text{robust EKF}} (\boldsymbol{\theta}_t) & = \frac{1}{2P_{n_t}} \mathbb{E}_{o_t \sim p(\cdot)} [\big( \delta^{\text{robust}}_{t, \boldsymbol{\theta}'}(\boldsymbol{\theta}_t, o_t) \big)^2] \\
	\nonumber & + \frac{1}{2}(\boldsymbol{\theta}_t - \boldsymbol{\hat{\theta}}_{t|t-1})^\top {\bf P}_{t|t-1}^{-1} (\boldsymbol{\theta}_t - \boldsymbol{\hat{\theta}}_{t|t-1}),
	\end{align*}
	where 
	\begin{equation}
	\label{eq:kalman update_robust}
	\boldsymbol{\hat{\theta}}^{\text{robust}}_{t|t} = \boldsymbol{\hat{\theta}}_{t|t-1} + \mathbb{E}_{o_t \sim p(\cdot)} \big[ {\bf K}_t \big( y^{\text{robust}}_{t, \boldsymbol{\theta}'} (o_t) - Q_{}(s_t, a_t; \boldsymbol{\hat{\theta}}_{t|t-1}) \big) \big],
	\end{equation}
	and $\boldsymbol{\hat{\theta}}^{\text{robust}}_{t|t} \in \arg\min_{\boldsymbol{\theta}_t} L_t^{\text{robust EKF}} (\boldsymbol{\theta}_t)$.}

\begin{proof}
	The proof for Theorem \ref{theorem2} follows the same arguments as the proof for Theorem \ref{theorem1}. The difference is in the target label definition. Here we use $y_{t, \boldsymbol{\theta}'}^{\text{robust}}$ instead of $y_{t, \boldsymbol{\theta}'}^{\text{nominal}}$. Since $y_{t, \boldsymbol{\theta}'}^{\text{robust}}$ does not depend on the random variable $\boldsymbol{\theta}_t$, it is fixed when taking the derivative of the objective function with respect to $\boldsymbol{\theta}_t$.
\end{proof}

\section{Cart-Pole experiment}
We describe here technical details regarding the Cart-Pole experiment. All algorithms (Double-DQN \cite{van2016deep}, RTD-DQN and Deep-RoK) used a feed forward fully connected neural network with 2 hidden layers with the tanh activation function. The input dimension is 4 $(x, \dot{x}, \theta, \dot{\theta})$, the output dimension is 2 (two possible actions) and each hidden layer is composed of $20$ neurons. The total number of weights in the network is $n=562$. The networks were trained on mini-batches of transitions, 10 samples in each mini-batch. We used a discount factor of $\gamma=0.9$. All algorithms were trained for 700 episodes. Double-DQN and RTD-DQN were trained with the Adam optimizer \cite{kingma2014adam} with learning rate $\alpha=0.001$. Deep-RoK was trained using the EKF (as described in the main paper) with learning rate $\alpha=1$, observation noise $P_{n_t}=0.001$, evolution noise ${\bf P}_{{\bf v}_t} = 0.01 \cdot {\bf I}$ and error covariance prior ${\bf P}_{0|0}={\bf I}$ ({\bf I} is the identity matrix). The results presented in the paper are the mean and standard deviation of the cumulative reward, obtained by applying a $\epsilon$- greedy policy ($\epsilon=0.1$) over the learned network for 500 episodes. 

The Double-DQN algorithm was trained on the nominal values {\it pole length}=$0.5$m and {\it cart mass}=$1.5$kg. 
The uncertainty set $\mathcal{P}(\psi), \psi=$ \{pole length, cart mass\} used in RTD-DQN and Deep-RoK algorithms was constructed by sampling uniformly values from the range:\\
pole length $\in \{ 0.2, 1.4\}$m;\\
cart mass $\in \{0.1, 7\}$kg. \\  
At the beginning of each episodes, we sampled 5 values from the above ranges and constructed the uncertainty set $\mathcal{P}$ for this episode.

\end{document}